\documentclass[11pt]{article}

\usepackage[margin=1in]{geometry}

\PassOptionsToPackage{numbers,compress}{natbib}
\usepackage{natbib}

\usepackage[utf8]{inputenc}
\usepackage[T1]{fontenc}
\usepackage{hyperref}
\usepackage{url}
\usepackage{booktabs}
\usepackage{amsfonts}
\usepackage{nicefrac}
\usepackage{microtype}
\usepackage{xcolor}

\usepackage{graphicx}
\usepackage{subcaption}
\usepackage{amsmath}
\usepackage{amssymb}
\usepackage{mathtools}
\usepackage{amsthm}
\usepackage[capitalize,noabbrev]{cleveref}
\usepackage{algorithm}
\usepackage{algorithmic}
\usepackage[textsize=tiny]{todonotes}
\usepackage{placeins}

\theoremstyle{plain}

\theoremstyle{definition}

\theoremstyle{remark}

\title{
Activation-Aware Weight Tensorization:
A Calibration-Time Preconditioner for Tensor-Network LLM Compression
}

\author{
Alessandro Beatini \\
Sapienza University of Rome \\
\texttt{alessandro.beatini@uniroma1.it}
\and
Marco Maronese \\
Independent Researcher \\
\texttt{marco.maronese95@gmail.com} \\
\href{https://orcid.org/0000-0001-5548-5947}
     {ORCID: 0000-0001-5548-5947}
\and
Emanuele Rodol\`a \\
Sapienza University of Rome \\
\texttt{rodola@di.uniroma1.it}
}

\date{}

\date{}

\begin{document}

\maketitle

\begin{abstract}
Post-training tensor-network compression replaces Transformer linear layers with
Tensor Train (TT) or Tree Tensor Network (TTN) operators, but standard
decompositions minimize weight-space Frobenius error rather than functional error
under the layer's activation distribution. We propose
\emph{Activation-aware Weight Tensorization} (AWT), a training-free calibration
wrapper that preconditions each weight matrix with a diagonal activation-derived
scale before an unchanged TT/TTN solver and deploys the result with only an
input-side elementwise rescaling. Across Llama~3.1~8B, Ministral~8B, and
Qwen2.5~7B, AWT consistently improves vanilla TT/TTN tensorization at
2--6$\times$ compression: under single-operator replacement, AWT closes 12--35\% of the WikiText
perplexity gap to the dense baseline across the three model families and
2--6$\times$ compression settings; while under
multi-operator Llama suffix replacement it closes 27--60\% across attention-group
and all-seven-matrix settings. The gains also transfer to downstream HellaSwag
and ARC-Challenge evaluations. We further show that diagonal preconditioning is
a robustness--modularity tradeoff rather than a diagonal-covariance assumption:
a dense full-covariance oracle wins its own weighted objective in 80/81 cases,
yet diagonal AWT gives better held-out functional fidelity in 53/81 cases.
Together, these results position AWT as a principled, modular preconditioner for
improving functional fidelity in fixed TT/TTN compression pipelines without
modifying the decomposition solver.
\end{abstract}

\section{Introduction}
Large language models (LLMs) are increasingly expensive to serve because dense
linear operators dominate memory movement and compute during inference
\citep{brown2020language,touvron2023llama,chowdhery2023palm,pope2023efficiently}.
This has motivated a broad set of post-training compression techniques,
including quantization, pruning, distillation, low-rank factorization, and
structured tensor decompositions \citep{zhu2023survey}.
Tensor-network compression offers a structured alternative to standard matrix
factorizations. A dense matrix $W\in\mathbb{R}^{m\times n}$ can be reshaped into
a higher-order tensor and represented by a network of low-rank cores: a Tensor
Train (TT) or matrix-product operator uses a chain of cores
\citep{oseledets2011tensor,novikov2015tensorizing}, while a Tree Tensor Network
(TTN) uses a loop-free hierarchy over tensor modes
\citep{shi2006classical,grasedyck2010hierarchical}. Unlike truncated SVD, which
imposes a single global rank on a matrix, tensor networks distribute rank
budgets across tensor modes and internal edges of a chosen topology. This richer
design space has made tensorization an increasingly relevant direction for LLM
compression, with recent methods and systems such as CompactifAI, TensorLLM, and
Minima demonstrating its use for compressing large neural operators and LLM
components
\citep{tomut2024compactifai,fovet2025accuracy,alvarez2025scaling,gu2025tensorllm,kozyrev2026practical}.

Despite this progress, post-training tensorization is typically activation-blind.
Once a tensorization shape, topology, and rank budget are chosen, standard TT/TTN
backends approximate the weight matrix by minimizing an isotropic reconstruction
objective in weight space. This objective is computationally convenient, but it
does not directly capture the error induced by replacing a layer inside a
pretrained model. 

This observation is well established in post-training quantization. Methods such
as SmoothQuant and AWQ show that simple calibration statistics can identify
activation channels whose perturbation disproportionately affects model behavior
\citep{xiao2023smoothquant,lin2024awq}. However, the same principle has not been
systematically exploited for post-training tensor-network decomposition. Tensor
decompositions introduce a different constraint from quantization: the
approximation is determined by a fixed tensorization layout, network topology,
and internal rank budget rather than by a bit-width. This raises a natural
question: 
can activation statistics be used to make existing tensor-network
solvers more function-preserving, without changing the solver itself?

We answer this question with \textbf{Activation-aware Weight Tensorization
(AWT)}. AWT computes a diagonal scale $D_\alpha$ from cached layer inputs,
decomposes the reparameterized matrix $W'=WD_\alpha$ using the same
tensor-network backend, tensorization layout, topology, and rank budget as
vanilla tensorization, and deploys the resulting approximation as $\widehat y = \widetilde W'(D_\alpha^{-1}x)$.
Thus, activation information changes the matrix presented to the decomposition
solver, but not the solver itself. The method is intentionally minimal: it
transfers the equivalent reparameterization idea from activation-aware
quantization to post-training tensor-network compression, requiring no
retraining, no custom TT/TTN solver, and no change to the tensorized operator
apart from a lightweight input-side diagonal scaling.
Empirically, we find that this simple preconditioning consistently improves the
functional fidelity of tensorized LLM operators. The gains persist across model
families, operator types, compression ratios, and tensor-network topologies, and
become especially important when multiple operators are compressed jointly. In
multi-operator suffix replacement, AWT closes 27--60\% of the WikiText
perplexity gap across both attention-group and all-seven-matrix settings
(\cref{tab:multiop_abs}), indicating that activation-aware preconditioning helps
control the accumulation of tensorization error inside the model.

\textbf{Contributions.}
Our main contributions are:
\emph{(i)} we introduce AWT, a diagonal activation-aware preconditioner for
post-training TT/TTN compression that preserves the underlying tensor-network
solver and deployment path;
\emph{(ii)} we show that AWT is equivalent to a column-weighted reconstruction
proxy (\cref{eq:awt_colweighted}), connecting tensorization error to
activation-conditioned functional distortion;
\emph{(iii)} we demonstrate consistent gains over vanilla tensorization in
single-operator and multi-operator LLM replacement experiments, with the largest
absolute gains arising when compression errors accumulate across operators
(\cref{tab:cross_model_ppl});
\emph{(iv)} we evaluate AWT across Llama~3.1~8B, Ministral~8B, and Qwen2.5~7B,
three compression ratios, seven operator types, and TT/TTN backends, including
matched-compression SVD and downstream task checks;
\emph{(v)} we compare diagonal AWT with a dense full-covariance oracle and show
that the diagonal restriction is a robustness--modularity tradeoff rather than a
diagonal-covariance assumption.

\section{Related Work}

\textbf{Low-rank and tensorized neural compression.}
Low-rank factorization has long been used to compress neural networks, from
matrix and filter decompositions for convolutional models to post-compression
fine-tuning
\citep{Denton2014,Jaderberg2014,Lebedev2015,Kim2015}. Tensor-network
parameterizations extend this idea by replacing dense operators with structured
multilinear factors. Tensor Train (TT) layers represent a matrix as a chain of
low-order cores, building on the TT format from numerical linear algebra
\citep{oseledets2011tensor,novikov2015tensorizing}; related formats such as
Tensor Rings and Block-Term decompositions have also been used for neural
compression \citep{Zhao2016,Li2017BTNets,Wang2018TRNets}. These methods provide
richer rank allocation than a single global matrix rank, but are typically
optimized using weight-space reconstruction objectives.

\textbf{Tensorization for language models.}
Tensorized representations have been explored in NLP for large embedding tables
and Transformer components, including TT-parameterized embeddings, tensorized
Transformer layers, and tensor-compressed models trained or distilled directly
\citep{hrinchuk2019tensorized,ma2019tensorized,Yang2023QATensorCompressed}.
Recent work has moved toward post-training tensorization of pretrained LLMs:
TensorGPT studies training-free TT/MPS compression of token embeddings
\citep{xu2023tensorgpt}; CompactifAI uses quantum-inspired tensor networks,
including MPO/TT-matrix-style decompositions, to compress attention and MLP
operators with recovery \citep{tomut2024compactifai,fovet2025accuracy,alvarez2025scaling};
TensorLLM tensorizes multi-head attention with Tucker-style structure
\citep{gu2025tensorllm}; and Minima presents a production-oriented pipeline
combining sensitivity prediction, Tucker/TT/TR decompositions, healing, and
custom kernels \citep{kozyrev2026practical}. These works establish tensorization
as an active LLM-compression direction. AWT is orthogonal: rather than proposing
a new topology or recovery procedure, it introduces a calibration-time
activation-aware preconditioner before standard TT/TTN decomposition.

\textbf{Function-aware compression objectives.}
A central limitation of weight-space compression is that Frobenius reconstruction
error can be misaligned with the functional error induced inside a pretrained
model. Fisher-weighted and distribution-aware decompositions address this by
weighting the objective with curvature, Fisher information, or input covariance
statistics \citep{hsu2022language,kalle2025distributionaware}. These methods
more directly target output distortion, but often require modified optimization,
dense preconditioners, or solver-specific updates. AWT instead uses activation
statistics only to construct a diagonal equivalent reparameterization, leaving
the tensor-network solver unchanged.

\textbf{Activation-aware post-training compression.}
Activation-aware scaling has been especially successful in post-training
quantization. SmoothQuant applies an equivalent rescaling that transfers
quantization difficulty from activation outliers into weights
\citep{xiao2023smoothquant}, while AWQ uses activation statistics to identify
salient channels for low-bit weight-only quantization \citep{lin2024awq}.
Related outlier-suppression methods also use equivalent transformations and
clipping to reduce structured activation outliers
\citep{Wei2022OutlierSuppression}. AWT brings this activation-aware perspective
to tensor-network decomposition, where the approximation budget is not a
bit-width but a fixed tensorization shape, topology, and rank budget.

\section{Background}

\textbf{Tensorizing linear operators.}
We consider a linear layer $y=Wx$ with
$W\in\mathbb{R}^{m\times n}$. Tensor-network compression begins by factorizing
the input and output dimensions as
$m=\prod_{k=1}^K m_k$ and $n=\prod_{k=1}^K n_k$, and reshaping $W$ into an
order-$2K$ tensor
$\mathcal{W}
\in
\mathbb{R}^{m_1\times\cdots\times m_K
\times n_1\times\cdots\times n_K}$.
The indices associated with $m_k$ correspond to output modes, while those
associated with $n_k$ correspond to input modes. A tensor-network approximation
then replaces $\mathcal{W}$ by a contraction of smaller core tensors whose
internal bond dimensions control the parameter budget.

\textbf{Tensor Train and Tree Tensor Network decompositions.}
The Tensor Train (TT) format, equivalently a Matrix Product Operator (MPO) for
linear maps, represents $\mathcal{W}$ as a chain of cores
$\{\mathcal{G}^{(k)}\}_{k=1}^K$, where
$\mathcal{G}^{(k)}\in
\mathbb{R}^{r_{k-1}\times m_k\times n_k\times r_k}$ and
$r_0=r_K=1$ \citep{oseledets2011tensor,novikov2015tensorizing}. Its entries are
given by
\begin{equation}
\mathcal{W}_{i_1,\dots,i_K,j_1,\dots,j_K}
=
\sum_{\alpha_1,\dots,\alpha_{K-1}}
\prod_{k=1}^K
\mathcal{G}^{(k)}_{\alpha_{k-1},i_k,j_k,\alpha_k}.
\label{eq:tt_mpo}
\end{equation}
The TT ranks $\{r_k\}$ are the internal bond dimensions and determine the
compression--expressivity trade-off. In practice, TT decompositions are commonly
computed by successive truncated SVDs of tensor unfoldings.
Tree Tensor Networks (TTNs) generalize this chain topology to a loop-free
hierarchical graph over tensor modes
\citep{shi2006classical,grasedyck2010hierarchical}. The leaves correspond to the
physical modes of $\mathcal{W}$, while internal nodes store latent cores that
merge groups of modes. Each internal edge $e$ carries a bond dimension
$\chi_e$, often bounded by a maximum rank $\chi$. Compared with TT, a TTN
allocates rank across a hierarchy rather than along a single ordering, yielding
a different inductive bias and parameter allocation for the same tensorized
operator.

\textbf{Activation-conditioned reconstruction.}
Standard post-training TT/TTN decompositions approximate $W$ by minimizing a
weight-space objective such as $\|W-\widetilde W\|_F^2$. However, replacing a
linear layer inside a pretrained model induces output error that depends on the
layer input distribution. For $y=Wx$ and approximation $\widehat W$,
\begin{equation}
\mathbb{E}\|(W-\widehat W)x\|_2^2
=
\|(W-\widehat W)\Sigma_x^{1/2}\|_F^2,
\qquad
\Sigma_x=\mathbb{E}[xx^\top].
\label{eq:functional_error_background}
\end{equation}
Thus, directions that are frequently or strongly activated should contribute
more to the reconstruction objective. Full covariance-weighted decompositions
can directly target this metric, but they generally require modified solvers or
dense preconditioners
\citep{hsu2022language,kalle2025distributionaware}.
AWT instead uses an equivalent diagonal reparameterization. For any invertible
diagonal matrix $D$,
\begin{equation}
Wx = (WD)(D^{-1}x),
\label{eq:equiv_scaling}
\end{equation}
so one can rescale the columns of $W$ before decomposition and compensate by
the inverse scaling at inference time. This is the same functional-preserving
principle used by activation-aware quantization methods such as SmoothQuant and
AWQ \citep{xiao2023smoothquant,lin2024awq}, but applied here to
tensor-network decomposition rather than bit-width allocation.

\section{Activation-aware Weight Tensorization (AWT)}
AWT is a calibration-time preconditioning wrapper for tensor-network compression
of linear layers. Given a linear map $y=Wx$ and a tensor-network backend
(TT/TTN) with a fixed tensorization layout, topology, and rank/parameter budget,
we compute a diagonal importance scaling $D_{\alpha}$ from cached input
activations and use it to bias the decomposition toward input channels that
matter more under typical inputs. Concretely, AWT (i) right-preconditions the
weight matrix by scaling its input-channel columns, (ii) tensorizes and
decomposes the scaled operator using a standard Frobenius-norm TT/TTN routine,
and (iii) deploys the compressed layer without materializing the dense unscaled
matrix: the inverse diagonal scaling is applied to the layer input, followed by
the TT/TTN operator (\cref{eq:awt_inference,alg:awt}).
As discussed in the background, the expected functional distortion of a compressed
linear layer induces a covariance-weighted metric
$\|(W-\widehat W)\Sigma^{1/2}\|_F$. Directly optimizing this full-metric
objective within TT/TTN would require either a solver that handles a
non-isotropic least-squares metric or an explicit dense preconditioner. The
latter globally mixes input channels, can be poorly aligned with a fixed
tensorization shape and rank budget, and would require a dense transform at
deployment. AWT instead uses a diagonal preconditioner. This restriction is
deliberate: it preserves the original channel layout seen by the tensorization,
requires only single-pass gradient-free calibration statistics, and leaves the
TT/TTN solver and contraction path unchanged. Thus, AWT trades covariance
optimality for modularity, statistical robustness, and compatibility with
off-the-shelf tensor-network decomposition.

\textbf{Key identity: diagonal preconditioning yields a weighted Frobenius proxy.}
Let $D\succ0$ be diagonal, define the preconditioned matrix $W'=WD$, and let
$\widetilde W'$ be any approximation to $W'$ returned by an off-the-shelf
Frobenius-norm TT/TTN routine after tensorization. Define the effective deployed
operator as $\widehat W=\widetilde W'D^{-1}$. Then
\begin{equation}
(W-\widehat W)D = WD-\widetilde W', \hspace{10mm}
\|(W-\widehat W)D\|_F = \|WD-\widetilde W'\|_F .
\label{eq:awt_identity}
\end{equation}
Equivalently, AWT minimizes a column-weighted reconstruction proxy,
\begin{equation}
\begin{aligned}
\|(W-\widehat W)D\|_F^2
&= \sum_{i=1}^{n} d_i^2
   \,\|W_{:i} - \widehat W_{:i}\|_2^2 ,
\end{aligned}
\label{eq:awt_colweighted}
\end{equation}
so columns associated with larger $d_i$ are fit more accurately. This is the
precise sense in which AWT injects activation information while reusing an
off-the-shelf TT/TTN Frobenius solver.

\textbf{Diagonal scales from activation importance.}
Given calibration activations $\{\mathbf{x}^{(t)}\}_{t=1}^T$ for a layer, we
compute per-channel importance scores $\{\mathrm{imp}_i\}_{i=1}^n$ and normalize
them by their mean
\(
\overline{\mathrm{imp}}=\frac{1}{n}\sum_{i=1}^n \mathrm{imp}_i.
\)
For numerical stability, we clamp the mean as
\(
\overline{\mathrm{imp}}_{\varepsilon}=\max(\overline{\mathrm{imp}},\varepsilon)
\)
and define scales $s_i$ via
\begin{equation}
s_i=
\max\!\left(
s_{\min}, 
\left(\frac{\mathrm{imp}_i}{\overline{\mathrm{imp}}_{\varepsilon}}\right)^{\alpha}
\right), \hspace{10mm}
s_{\min} =
\max(\texttt{min\_scale},\varepsilon).
\label{eq:awt_scales}
\end{equation}
where $\alpha>0$ controls the strength of reweighting, $\varepsilon$ is a small
numerical constant, and $s_{\min}$ prevents large factors in $D_{\alpha}^{-1}$.
We set $D_{\alpha}=\mathrm{diag}(s_1,\dots,s_n)$. In our experiments, we
instantiate $\mathrm{imp}_i$ using either MeanAbs,
$\mathrm{imp}_i^{\text{MeanAbs}}=\mathbb{E}_t[|x_i^{(t)}|]$, or Std,
$\mathrm{imp}_i^{\text{Std}}=\sqrt{\mathrm{Var}_t(x_i^{(t)})+\varepsilon}$,
with statistics taken over cached calibration activations.
We use MeanAbs and Std because they are single-pass, gradient-free statistics
that can be estimated from cached calibration activations, preserving the
training-free character of AWT. Gradient- or Fisher-based saliency measures are
possible, but would require backpropagation and would change the computational
profile of the method.
The diagonal restriction should not be interpreted as assuming diagonal
activation covariance; rather, it preserves the original channel layout and
keeps the preconditioner statistically and computationally lightweight.

\textbf{Precondition--tensorize--deploy (without materializing $\widehat W$).}
For input-channel scaling (right preconditioning), we scale the columns of $W$ and tensorize the scaled operator:
\begin{equation}
W' = W D_{\alpha},
\label{eq:awt_precond}
\end{equation}
apply a tensor-network decomposition to obtain $\widetilde{W}' \approx W'$, and conceptually define
\begin{equation}
\widehat{W} = \widetilde{W}' D_{\alpha}^{-1}.
\label{eq:awt_unscale}
\end{equation}
In practice, to avoid materializing $\widehat{W}$ (which would be dense), we implement the layer as the
\emph{equivalent} computation
\begin{equation}
\widehat{y} = \widehat{W}x = (\widetilde{W}'D_{\alpha}^{-1})x = \widetilde{W}'(D_{\alpha}^{-1}x),
\label{eq:awt_inference}
\end{equation}
i.e., we apply the inverse diagonal scaling to the layer input and then apply the TT/TTN-compressed operator
$\widetilde{W}'$. Since $D_{\alpha}$ is diagonal, this adds only an elementwise per-channel multiply and can be
fused with surrounding kernels. This preserves the TT/TTN parameter budget while remaining functionally
equivalent to Eq.~\eqref{eq:awt_unscale}.
The algorithm is intentionally modular: any TT/TTN decomposition implementation can be used in the decomposition step, and the only data-aware component is the diagonal scaling computed from a small calibration set.

\begin{algorithm}[t]
\caption{Activation-aware Weight Tensorization (AWT)}
\label{alg:awt}
\begin{algorithmic}[1]
\REQUIRE Weight matrix $W \in \mathbb{R}^{m \times n}$, calibration inputs $\mathcal{C}$,
importance statistic $\mathrm{imp}(\cdot)$ (MeanAbs or Std), exponent $\alpha>0$,
numerical constant $\varepsilon>0$, minimum scale \texttt{min\_scale}$>0$,
tensorization shape, tensor-network routine $\textsc{Decomp}(\cdot)$ (TT/TTN)
with target ranks/budget.
\ENSURE Compressed operator implemented as $(\widetilde{W}', D_{\alpha}^{-1})$.
\STATE Run the model on $\mathcal{C}$ and collect layer input activations $\{\mathbf{x}^{(t)}\}_{t=1}^T$.
\STATE Compute per-channel importances $\mathrm{imp}_i \leftarrow \mathrm{imp}(\{x_i^{(t)}\}_{t=1}^T)$ for $i=1,\dots,n$.
\STATE Compute mean importance $\overline{\mathrm{imp}} \leftarrow \frac{1}{n}\sum_{i=1}^n \mathrm{imp}_i$
and clamp $\overline{\mathrm{imp}}_{\varepsilon} \leftarrow \max(\overline{\mathrm{imp}},\varepsilon)$.
\STATE Set $s_{\min} \leftarrow \max(\texttt{min\_scale},\varepsilon)$ and form scales
$s_i \leftarrow \max\!\left(s_{\min},
\left(\frac{\mathrm{imp}_i}{\overline{\mathrm{imp}}_{\varepsilon}}\right)^{\alpha}\right)$.
\STATE Form $D_{\alpha} \leftarrow \mathrm{diag}(s_1,\dots,s_n)$.
\STATE Precondition (right scaling): $W' \leftarrow W D_{\alpha}$.
\STATE Tensorize/reshape $W'$ into $\mathcal{W}'$ using the specified shape.
\STATE Decompose: $\widetilde{\mathcal{W}}' \leftarrow \textsc{Decomp}(\mathcal{W}')$ (subject to target ranks/budget).
\STATE Untensorize $\widetilde{\mathcal{W}}'$ into a (compressed) operator $\widetilde{W}'$.
\STATE \textbf{Deploy without materializing $\widehat W$:} implement $\widehat y = \widetilde{W}'(D_{\alpha}^{-1}x)$.
\RETURN $(\widetilde{W}', D_{\alpha}^{-1})$.
\end{algorithmic}
\end{algorithm}

\section{Experimental Validation}

\textbf{Models and scope.}
We evaluate AWT on Llama~3.1~8B, Ministral~8B, and Qwen2.5~7B, considering the
seven dense Transformer operators
$\{W_q,W_k,W_v,W_o,W_{up},W_{gate},W_{down}\}$ at
$2/4/6\times$ compression. Unless stated otherwise, vanilla and AWT use the
same tensorization layout, topology, rank budget, calibration set, and TT/TTN
backend; only the diagonal preconditioning differs.
Our experiments are organized around four questions:
(i) whether activation-aware preconditioning improves the local functional
fidelity of a fixed TT/TTN approximation;
(ii) whether these local improvements translate into model-level behavior under
single-operator replacement;
(iii) how the gains behave when multiple compressed operators are composed; and
(iv) how the diagonal AWT proxy compares with alternative baselines, including
matched-compression SVD and dense full-covariance preconditioning.

\textbf{Calibration data and activation caches.}
AWT estimates per-channel activation statistics from a small unlabeled
calibration set. We draw calibration prompts from a subset of lm-eval tasks
(MMLU, HellaSwag, and BoolQ), run the original dense model, and record the input
activations to each targeted linear operator using forward hooks. Only the input
vectors immediately before multiplication by $W$ are cached; labels and task
losses are never used. Calibration examples are kept disjoint from all examples
used for metric caches and downstream evaluation.
For layer-level evaluation, we construct a separate held-out \emph{metric cache}
of input activations $X$. This cache is used only to measure local output error
and to select among candidate tensorized configurations at a fixed parameter
budget. Separating calibration and metric caches avoids selecting tensorized
operators on the same activations used to construct the AWT scales.

\subsection{Compared methods}

\textbf{Tensorization backends.}
We compare Tensor Train (\textbf{TT}) and two Tree Tensor Network variants
(\textbf{TTN4}, \textbf{TTN8}), which differ in tree depth/fanout and therefore
impose different rank-allocation biases. For a given matrix and compression
budget, vanilla tensorization and AWT use the same tensorization shape, mode
ordering, topology, and rank budget. This fixed-layout protocol isolates the
effect of activation-aware preconditioning rather than jointly optimizing the
tensorization search space.

\textbf{Activation-aware preconditioning.}
For each TT/TTN backend, we evaluate \textbf{Vanilla} tensorization with no
preconditioning, \textbf{AWT-MeanAbs} with channel scales from
$\mathbb{E}[|x_i|]$, and \textbf{AWT-Std} with channel scales from the
per-channel standard deviation. All AWT variants use right preconditioning
(input-channel scaling) and deploy the compressed operator as
$\widetilde W'(D_\alpha^{-1}x)$. We sweep
$\alpha\in\{0.5,1.0,1.5\}$ and use $\alpha=1.0$ as the default unless otherwise
specified.

\textbf{Baselines and oracles.}
We include matched-compression truncated SVD as a standard matrix low-rank
baseline under the same single-operator protocol. We also compare diagonal AWT
with a dense full-covariance preconditioner in an offline operator-level
analysis. The full-covariance variant is used as an oracle-style comparison for
the richer covariance-weighted objective, while diagonal AWT is the practical
plug-in method used in deployment.

\textbf{Compression matching.}
To compare methods at equal compression, we sweep TT/TTN rank configurations and
select, for each target matrix, candidates satisfying the specified compression
ratio. Unless otherwise stated, parameter counts include the tensor-network
cores; the additional AWT diagonal scale is one input-channel vector and is
negligible relative to the compressed operator, but we report it explicitly in
the implementation details. Candidates are ranked using the held-out
layer-level functional fidelity metric defined below.

\textbf{Layer-level metrics.}
Let $X\in\mathbb{R}^{B\times n}$ be a held-out activation cache for a layer with
weight matrix $W\in\mathbb{R}^{m\times n}$, and let
$Y=XW^\top$. For AWT, predictions are computed without materializing
$\widehat W$ as
$\widehat Y=(XD_\alpha^{-1})(\widetilde W')^\top$, where
$\widetilde W'$ approximates $W'=WD_\alpha$. Our primary local metric is
activation-conditioned relative output error,


\begin{equation}
    \mathrm{RelFrob}_{out}
=
\frac{\|\widehat Y-Y\|_F}{\|Y\|_F+\varepsilon},
\qquad \varepsilon=10^{-12}.
\end{equation}

For comparison, we also report weight-space reconstruction error
$\mathrm{RelFrob}_{w}=\|W-\widehat W\|_F/(\|W\|_F+\varepsilon)$.
For the full-covariance oracle analysis only, we use
$\mathrm{RelFrob}_{\Sigma}
=
\|(W-\widehat W)\Sigma_X^{1/2}\|_F/
(\|W\Sigma_X^{1/2}\|_F+\varepsilon)$.
\vspace{-0.7em}
\nopagebreak[4]

\subsection{Evaluation protocols}
\vspace{-0.4em}
\textbf{Local layer-wise sweep.}
For each targeted matrix instance, backend, compression ratio, and AWT setting,
we sweep admissible TT/TTN rank configurations and compute local metrics on the
held-out metric cache. This sweep is performed on activations from the original
dense model and is independent of end-to-end benchmark scores. At each parameter
budget, we retain the candidate with the lowest
$\mathrm{RelFrob}_{\mathrm{out}}$ unless otherwise stated. This protocol gives a
controlled comparison between vanilla tensorization and AWT under the same
tensor-network hypothesis class.

\textbf{Single-operator replacement.}
To test whether local improvements translate into model-level behavior, we
replace one dense operator at a time with its selected tensorized approximation
and evaluate the resulting model. Results are aggregated over blocks, operator
types, and backends at each compression ratio. This protocol isolates the effect
of compressing a single matrix while preserving the surrounding model.

\textbf{Multi-operator suffix replacement.}
Single-operator replacement is a controlled diagnostic, but deployed compression
typically composes errors from multiple approximated operators. We therefore
also evaluate suffix replacement on Llama~3.1~8B: for the last $k$ Transformer
blocks, we jointly replace either the attention group
$\{W_q,W_k,W_v,W_o\}$ or all seven dense operators. This evaluates whether the
AWT-vs-vanilla advantage persists when tensorization errors accumulate through
multiple compressed operators.

\textbf{End-to-end metrics.}
We report WikiText perplexity for both single-operator and multi-operator
replacement. We additionally evaluate downstream accuracy on HellaSwag and
ARC-Challenge as external task-level checks. These downstream results are used
as corroborating evidence rather than as the primary selection criterion; no
candidate is selected using downstream task accuracy.

\textbf{Runtime and deployment overhead.}
AWT performs calibration and scale construction offline. At inference time, the
only additional operation relative to the underlying tensorized layer is the
input-side elementwise multiplication $D_\alpha^{-1}x$ before the TT/TTN
contraction. The tensorized operator and contraction path are unchanged, and the
scaling can be fused into surrounding kernels. We therefore separate AWT's
preconditioning overhead from the latency properties of a particular optimized
TT/TTN backend.

\section{Results}

Across local, single-operator, and multi-operator evaluations, AWT consistently
improves vanilla TT/TTN tensorization at matched parameter budgets. The local
effect appears in activation-conditioned output error, transfers to WikiText
perplexity across three model families, and becomes more pronounced when several
operators are compressed jointly. We also find that truncated SVD remains a
strong matrix low-rank baseline at matched parameter count, while dense
full-covariance preconditioning is not a uniformly better practical substitute
for the diagonal AWT proxy.

\textbf{Local fidelity and metric alignment}
AWT first improves the quantity it is designed to affect: held-out
activation-conditioned output fidelity. At matched TT/TTN parameter budgets,
moderate activation reweighting reduces
$\mathrm{RelFrob}_{\mathrm{out}}$ relative to vanilla tensorization
(\cref{tab:local_metric_summary}). The optimum for functional fidelity differs
from the optimum for weight-space reconstruction: increasing $\alpha$ continues
to reduce $\mathrm{RelFrob}_{\mathrm{w}}$, but the best
$\mathrm{RelFrob}_{\mathrm{out}}$ occurs around $\alpha=1$. This separation
supports the view that preserving the action of a compressed layer on real
activations is not equivalent to minimizing plain Frobenius error in weight
space.
The same pattern appears at the model level. Across single-operator replacement
trials, changes in activation-conditioned output error are more predictive of
changes in perplexity than changes in weight-space error
(\cref{tab:local_metric_summary}, right). We therefore use held-out output
fidelity as the candidate-selection metric for end-to-end replacement.

\begin{table*}[t]
\caption{Local functional fidelity and metric alignment. \textbf{Left}: effect of the
AWT exponent $\alpha$ at $4\times$ compression on Llama~3.1~8B. AWT reduces
activation-conditioned output error, while weight-space error follows a
different trend. \textbf{Right}: Spearman correlation between changes in local metrics
and changes in WikiText perplexity under single-operator replacement
($\alpha=1.0$), grouped by operator family. Groups: QK$=\{W_q,W_k\}$, VO$=\{W_v,W_o\}$,
MLP$=\{W_{up},W_{down},W_{gate}\}$. Higher correlation means the local
metric better predicts model-level degradation.}
\label{tab:local_metric_summary}
\centering
\small
\begin{minipage}[t]{0.43\textwidth}
\centering
\textbf{AWT exponent ablation at $4\times$}\\[0.35em]
\setlength{\tabcolsep}{3pt}
\begin{tabular}{llrrrr}
\toprule
& & \multicolumn{2}{c}{Out.} & \multicolumn{2}{c}{Weight} \\
\cmidrule(lr){3-4}\cmidrule(lr){5-6}
$\alpha$ & Stat. & Val. & $\Delta$ & Val. & $\Delta$ \\
\midrule
0.5 & Std     & 0.606 & -9.2 & 0.754 & -1.7 \\
0.5 & MeanAbs & 0.594 & -10.9 & 0.751 & -2.0 \\
1.0 & Std     & 0.583 & -12.6 & 0.691 & -9.9 \\
1.0 & MeanAbs & 0.577 & -13.5 & 0.665 & -13.2 \\
1.5 & Std     & 0.589 & -11.6 & 0.526 & -31.4 \\
1.5 & MeanAbs & 0.592 & -11.3 & 0.455 & -40.6 \\
\bottomrule
\end{tabular}

\vspace{0.4em}
\footnotesize
$\Delta$ values are percentages.
\end{minipage}
\hfill
\begin{minipage}[t]{0.54\textwidth}
\centering
\textbf{Metric--PPL rank correlation}\\[0.35em]
\setlength{\tabcolsep}{3.5pt}
\begin{tabular}{llrrrr}
\toprule
Model & Metric & Overall & QK & VO & MLP \\
\midrule
Llama~3.1~8B
& $\Delta$Out & \textbf{0.427} & \textbf{0.680} & 0.219 & \textbf{0.373} \\
& $\Delta$Weight & 0.246 & -0.207 & -0.290 & 0.265 \\
\midrule
Ministral~8B
& $\Delta$Out & \textbf{0.395} & \textbf{0.794} & 0.245 & \textbf{0.235} \\
& $\Delta$Weight & 0.042 & -0.299 & 0.438 & -0.072 \\
\midrule
Qwen2.5~7B
& $\Delta$Out & \textbf{0.327} & \textbf{0.401} & \textbf{0.458} & \textbf{0.165} \\
& $\Delta$Weight & -0.049 & -0.213 & -0.260 & 0.164 \\
\bottomrule
\end{tabular}

\vspace{0.4em}
\footnotesize
QK $=\{W_q,W_k\}$, VO $=\{W_v,W_o\}$, MLP $=\{W_{up},W_{down},W_{gate}\}$.
$\Delta$Out denotes $\Delta\mathrm{RelFrob}_{\mathrm{out}}$ and
$\Delta$Weight denotes $\Delta\mathrm{RelFrob}_{\mathrm{w}}$.
\end{minipage}
\end{table*}

\textbf{Single-operator replacement across model families}
We next replace one dense operator at a time with its selected tensorized
approximation and evaluate WikiText perplexity. This protocol isolates the
effect of compressing one operator while preserving the rest of the model. As
shown in \cref{tab:cross_model_ppl}, AWT improves over vanilla tensorization at
every tested compression ratio across all three model families. The absolute
perplexity changes are small in this controlled setting, but the direction is
consistent and the fraction of the dense-to-vanilla gap recovered remains
substantial across architectures.

\begin{table*}[t]
\caption{Single-operator and multi-operator WikiText evaluation.
\textbf{Left:} single-operator replacement across model families; lower PPL is better.
``Gap closed'' is computed relative to the dense baseline of the corresponding model.
\textbf{Right:} multi-operator suffix replacement on Llama~3.1~8B; entries are
$\Delta$PPL $=$ PPL$_{\mathrm{AWT}}-$PPL$_{\mathrm{vanilla}}$, so negative is better.}
\label{tab:cross_model_ppl}
\label{tab:multiop_abs}
\centering
\small
\begin{minipage}[t]{0.55\textwidth}
\centering
\textbf{Single-operator replacement}\\[0.25em]
\setlength{\tabcolsep}{3.5pt}
\begin{tabular}{llcccc}
\toprule
Model & Comp. & Dense & Vanilla & AWT & Gap closed \\
\midrule
Llama~3.1~8B
& $2\times$ & 8.64 & 8.79 & 8.76 & 20.4\% \\
& $4\times$ & 8.64 & 8.90 & 8.85 & 17.1\% \\
& $6\times$ & 8.64 & 8.97 & 8.93 & 12.2\% \\
\midrule
Ministral~8B
& $2\times$ & 8.66 & 8.74 & 8.71 & 35.3\% \\
& $4\times$ & 8.66 & 8.81 & 8.76 & 31.3\% \\
& $6\times$ & 8.66 & 8.86 & 8.80 & 28.6\% \\
\midrule
Qwen2.5~7B
& $2\times$ & 9.59 & 9.76 & 9.70 & 33.5\% \\
& $4\times$ & 9.59 & 9.85 & 9.79 & 25.8\% \\
& $6\times$ & 9.59 & 9.91 & 9.84 & 22.2\% \\
\bottomrule
\end{tabular}
\end{minipage}
\hfill
\begin{minipage}[t]{0.40\textwidth}
\centering
\textbf{Multi-operator suffix replacement}\\[0.25em]
\setlength{\tabcolsep}{5pt}
\begin{tabular}{llccc}
\toprule
Scope & $k$ & $2\times$ & $4\times$ & $6\times$ \\
\midrule
Attention
& 1 & -0.41 & -0.47 & -0.46 \\
& 2 & -0.59 & -0.72 & -0.81 \\
& 4 & -0.76 & -1.02 & -1.11 \\
& 8 & -1.40 & -2.48 & -2.75 \\
\midrule
All seven
& 1 & -1.86 & -2.61 & -3.62 \\
& 2 & -7.74 & -9.21 & -11.71 \\
\bottomrule
\end{tabular}
\end{minipage}
\end{table*}

\textbf{Composition: multi-operator suffix replacement}
Single-operator replacement tests isolated substitutions; practical compression
composes several approximations. We therefore evaluate suffix replacement on
Llama~3.1~8B, jointly replacing either the attention group
$\{W_q,W_k,W_v,W_o\}$ or all seven dense operators in the last $k$ Transformer
blocks. \Cref{tab:multiop_abs} reports
$\Delta\mathrm{PPL}=\mathrm{PPL}_{\mathrm{AWT}}-
\mathrm{PPL}_{\mathrm{vanilla}}$, so negative values favor AWT.
AWT improves over vanilla tensorization in every tested suffix setting. The
absolute gain grows as more operators are replaced, indicating that
activation-aware preconditioning helps control the accumulation of tensorization
error through the model.

\subsection{Baselines, downstream tasks, and deployment overhead}

\textbf{Matched-compression SVD.}
We include truncated SVD as a strong matrix low-rank reference under the same
single-operator protocol. as detailed in Appendix \cref{tab:app_svd_cov}., A matched-compression SVD baseline is stronger than the TT/TTN pipeline under
the single-operator protocol, with WikiText PPL $8.71/8.81/8.87$ at
$2/4/6\times$, compared to $8.76/8.85/8.93$ for AWT and
$8.79/8.90/8.97$ for vanilla TT/TTN. This clarifies the role of AWT: it improves
a fixed tensor-network compression family rather than claiming that TT/TTN
universally dominates matrix low-rank compression.

\textbf{Downstream task validation.}
As an additional task-level check, we evaluate HellaSwag and ARC-Challenge under
single-operator replacement on Llama~3.1~8B. AWT improves over vanilla
tensorization at all tested compression ratios on both tasks. The absolute gains
are small, as expected in the controlled single-operator setting: HellaSwag
improves by $0.0003$--$0.0008$ accuracy points and ARC-Challenge improves by
$0.0009$--$0.0017$. We report the full backend-specific downstream breakdowns
in the appendix.

\textbf{Deployment overhead.}
AWT performs calibration, importance estimation, and scale construction offline.
At inference time, the only extra operation relative to the underlying
tensorized layer is the elementwise input scaling $D_\alpha^{-1}x$ before the
TT/TTN contraction. The tensorized operator and contraction path are unchanged,
and the scaling can typically be fused into surrounding kernels. We therefore
treat optimized TT/TTN latency as a property of the tensorization backend, with
AWT adding only a lightweight input-side rescaling.

\textbf{Diagonal-design analysis.}
The diagonal preconditioner should not be interpreted as a diagonal-covariance
assumption. Activations are strongly non-diagonal: the diagonal covariance mass
is only $0.032/0.142/0.212$ for Llama/Ministral/Qwen, while the off-diagonal
Frobenius ratio is $0.984/0.926/0.885$. A dense full-covariance oracle wins its
own covariance-weighted objective in $80/81$ cases, but diagonal AWT gives lower
held-out output error in $53/81$ cases. This supports diagonal AWT as a
robustness--modularity tradeoff; full details are reported in
\cref{tab:app_covariance}.

\section{Discussion and Conclusion}

Our results show that activation-aware preconditioning improves post-training
tensor-network compression of LLM linear operators. Across local fidelity,
single-operator replacement, and multi-operator suffix replacement, AWT
consistently improves vanilla TT/TTN tensorization at matched parameter budgets
across Llama~3.1~8B, Ministral~8B, and Qwen2.5~7B. The gains become especially
pronounced when multiple operators are replaced jointly, suggesting that
activation-aware preconditioning helps control accumulated tensorization error.

\textbf{Functional fidelity versus weight reconstruction.}
Weight-space reconstruction error is not the right proxy for the behavior of a
compressed layer inside an LLM. The activation-conditioned output metric is more
aligned with perplexity changes than plain Frobenius weight error, especially
for attention projections (\cref{tab:local_metric_summary}). This supports the
design of AWT: the preconditioner does not change the tensor-network hypothesis
class, but changes how a fixed rank budget is allocated by the solver, biasing
reconstruction toward input channels that matter under real activations. It also
explains why the best $\alpha$ for functional fidelity need not minimize
weight-space error.

\textbf{Topology- and operator-dependent tensorizability.}
Tensorizability is not uniform across operators or topologies. TTN backends
often outperform TT at the same parameter budget, with the clearest gap on
attention projections, particularly $W_q$ and $W_k$. A plausible explanation is
structural alignment: multi-head attention induces head-wise and block-structured
subspaces, and a hierarchical tree can merge correlated groups earlier than a
one-dimensional TT chain
\citep{vaswani2017attention,shi2006classical,grasedyck2010hierarchical,hackbusch2009h2}.
This remains an empirical hypothesis, since performance depends on
tensorization shape, mode ordering, rank allocation, and TTN topology. MLP
matrices are generally harder to tensorize at matched compression, consistent
with the view that feed-forward blocks implement richer feature mixing and
content storage \citep{geva2021ffn_kv}. These differences suggest that practical
tensorization pipelines should allocate ranks and choose topologies in an
operator-aware manner.

\textbf{Relationship to SVD and other compression families.}
The matched-compression SVD baseline highlights that standard matrix low-rank
approximation remains strong at matched parameter count. Tensor networks offer a
different structured compression family, with multi-core representations,
explicit internal rank budgets, and topology choices that may be useful for
deployment and for modeling structured correlations after tensorization. AWT
should therefore be viewed as improving this tensor-network family rather than
replacing other compression primitives. It is also complementary to quantization,
pruning, and matrix factorization: one could apply AWT before tensorizing
weights and then further quantize the resulting tensor-network cores.

\textbf{Diagonal preconditioning as a robustness--modularity tradeoff.}
The diagonal restriction in AWT should not be interpreted as assuming diagonal
activation covariance. Our covariance analysis shows the opposite: activations
are strongly non-diagonal. Dense full-covariance preconditioning optimizes its
intended covariance-weighted objective more directly, but does not reliably
improve held-out local output fidelity. Diagonal AWT gives up optimality on that
objective, but preserves the original channel layout, avoids dense deployment
transforms, reduces the statistical estimation burden, and leaves the TT/TTN
solver and contraction path unchanged.

\textbf{Limitations and future work.}
Several limitations remain. First, tensorization quality depends on design
choices that we keep fixed or only partially sweep, including dimension
factorization, mode ordering, rank allocation, and TTN topology. Jointly
optimizing these choices with activation-aware preconditioning is a natural next
step. Second, our runtime analysis isolates the overhead of AWT itself, an
input-side diagonal scaling, but end-to-end latency depends on optimized TT/TTN
kernels and hardware-specific implementations. Third, calibration statistics are
estimated from a small input set and may not capture all downstream
distributions. Finally, while our multi-operator suffix experiments move beyond
isolated substitutions, full-model compression requires coordinated choices
across layers and operators.

In summary, AWT is a training-free, solver-agnostic mechanism for making
post-training tensor-network compression more activation-aware. By inserting a
diagonal equivalent reparameterization before standard TT/TTN decomposition, it
improves functional fidelity without retraining, custom solvers, or changes to
the tensorized operator, especially when approximation errors accumulate across
multiple compressed operators.

\bibliographystyle{plainnat}
\bibliography{awt_references}

\newpage

\newpage
\FloatBarrier
\clearpage
\appendix

\makeatletter
\@addtoreset{table}{section}
\@addtoreset{figure}{section}
\makeatother
\renewcommand{\thetable}{\thesection.\arabic{table}}
\renewcommand{\thefigure}{\thesection.\arabic{figure}}

\section{Additional Experimental Results}
\label{app:additional_results}

This appendix provides additional experimental details and breakdowns omitted
from the main text for space: downstream task results, deployment overhead,
covariance analysis, cross-model ablations, backend comparisons, and
multi-operator replacement tables.

\FloatBarrier
\subsection{Llama per-operator output-error breakdowns}
\label{app:llama_per_operator}

\begin{table}[H]
\caption{Per-operator activation-conditioned output error (relative Frobenius; lower is better) under TTN depth4 tensorization with $\alpha=1$.
Entries are reported as mean $\mu$ and standard deviation $\sigma$ over $N=32$ single-operator trials per matrix type.}
\label{tab:perop_outerr_ttn4}
\centering
\footnotesize
\setlength{\tabcolsep}{3.2pt}
\renewcommand{\arraystretch}{1.12}
\resizebox{\textwidth}{!}{%
\begin{tabular}{lccc ccc ccc c}
\toprule
& \multicolumn{3}{c}{2$\times$ compression} & \multicolumn{3}{c}{4$\times$ compression} & \multicolumn{3}{c}{6$\times$ compression} & \\
\cmidrule(lr){2-4}\cmidrule(lr){5-7}\cmidrule(lr){8-10}
Matrix
& Vanilla ($\mu\!\pm\!\sigma$) & Std ($\mu\!\pm\!\sigma$) & MeanAbs ($\mu\!\pm\!\sigma$)
& Vanilla ($\mu\!\pm\!\sigma$) & Std ($\mu\!\pm\!\sigma$) & MeanAbs ($\mu\!\pm\!\sigma$)
& Vanilla ($\mu\!\pm\!\sigma$) & Std ($\mu\!\pm\!\sigma$) & MeanAbs ($\mu\!\pm\!\sigma$)
& $N$ \\
\midrule
$W_q$    & $0.180\pm0.047$ & $0.147\pm0.041$ & $0.147\pm0.041$
         & $0.245\pm0.056$ & $0.205\pm0.052$ & $0.205\pm0.052$
         & $0.279\pm0.060$ & $0.236\pm0.057$ & $0.236\pm0.057$ & 32 \\
$W_k$    & $0.192\pm0.049$ & $0.147\pm0.037$ & $0.147\pm0.037$
         & $0.304\pm0.067$ & $0.230\pm0.050$ & $0.230\pm0.050$
         & $0.358\pm0.070$ & $0.274\pm0.054$ & $0.274\pm0.053$ & 32 \\
$W_v$    & $0.595\pm0.065$ & $0.470\pm0.048$ & $0.467\pm0.047$
         & $0.749\pm0.061$ & $0.623\pm0.050$ & $0.621\pm0.049$
         & $0.811\pm0.054$ & $0.692\pm0.045$ & $0.689\pm0.045$ & 32 \\
$W_o$    & $0.526\pm0.087$ & $0.387\pm0.089$ & $0.389\pm0.085$
         & $0.665\pm0.098$ & $0.525\pm0.107$ & $0.526\pm0.104$
         & $0.722\pm0.100$ & $0.591\pm0.113$ & $0.591\pm0.110$ & 32 \\
$W_{up}$ & $0.597\pm0.078$ & $0.559\pm0.075$ & $0.561\pm0.076$
         & $0.646\pm0.084$ & $0.611\pm0.081$ & $0.613\pm0.082$
         & $0.698\pm0.091$ & $0.666\pm0.087$ & $0.668\pm0.088$ & 32 \\
$W_{down}$ & $0.677\pm0.094$ & $0.535\pm0.157$ & $0.547\pm0.158$
           & $0.730\pm0.098$ & $0.587\pm0.171$ & $0.600\pm0.172$
           & $0.784\pm0.101$ & $0.644\pm0.184$ & $0.656\pm0.186$ & 32 \\
$W_{gate}$ & $0.396\pm0.061$ & $0.374\pm0.058$ & $0.376\pm0.059$
           & $0.432\pm0.067$ & $0.411\pm0.063$ & $0.413\pm0.064$
           & $0.470\pm0.073$ & $0.452\pm0.069$ & $0.454\pm0.070$ & 32 \\
\bottomrule
\end{tabular}%
}
\end{table}

\begin{table}[H]
\caption{Per-operator activation-conditioned output error (relative Frobenius; lower is better) under TTN depth8 tensorization with $\alpha=1$.
Entries are reported as mean $\mu$ and standard deviation $\sigma$ over $N=32$ single-operator trials per matrix type.}
\label{tab:perop_outerr_ttn8}
\centering
\footnotesize
\setlength{\tabcolsep}{3.2pt}
\renewcommand{\arraystretch}{1.12}
\resizebox{\textwidth}{!}{%
\begin{tabular}{lccc ccc ccc c}
\toprule
& \multicolumn{3}{c}{2$\times$ compression} & \multicolumn{3}{c}{4$\times$ compression} & \multicolumn{3}{c}{6$\times$ compression} & \\
\cmidrule(lr){2-4}\cmidrule(lr){5-7}\cmidrule(lr){8-10}
Matrix
& Vanilla ($\mu\!\pm\!\sigma$) & Std ($\mu\!\pm\!\sigma$) & MeanAbs ($\mu\!\pm\!\sigma$)
& Vanilla ($\mu\!\pm\!\sigma$) & Std ($\mu\!\pm\!\sigma$) & MeanAbs ($\mu\!\pm\!\sigma$)
& Vanilla ($\mu\!\pm\!\sigma$) & Std ($\mu\!\pm\!\sigma$) & MeanAbs ($\mu\!\pm\!\sigma$)
& $N$ \\
\midrule
$W_q$    & $0.182\pm0.047$ & $0.148\pm0.041$ & $0.148\pm0.041$
         & $0.247\pm0.057$ & $0.207\pm0.052$ & $0.207\pm0.052$
         & $0.282\pm0.060$ & $0.239\pm0.057$ & $0.239\pm0.057$ & 32 \\
$W_k$    & $0.197\pm0.050$ & $0.151\pm0.038$ & $0.151\pm0.038$
         & $0.305\pm0.066$ & $0.233\pm0.051$ & $0.232\pm0.050$
         & $0.357\pm0.069$ & $0.275\pm0.053$ & $0.275\pm0.053$ & 32 \\
$W_v$    & $0.603\pm0.065$ & $0.478\pm0.048$ & $0.475\pm0.047$
         & $0.770\pm0.065$ & $0.627\pm0.048$ & $0.624\pm0.048$
         & $0.833\pm0.061$ & $0.691\pm0.045$ & $0.688\pm0.044$ & 32 \\
$W_o$    & $0.530\pm0.088$ & $0.390\pm0.090$ & $0.393\pm0.086$
         & $0.670\pm0.098$ & $0.530\pm0.108$ & $0.531\pm0.105$
         & $0.727\pm0.100$ & $0.597\pm0.113$ & $0.597\pm0.111$ & 32 \\
$W_{up}$ & $0.597\pm0.078$ & $0.559\pm0.075$ & $0.561\pm0.076$
         & $0.649\pm0.085$ & $0.613\pm0.082$ & $0.615\pm0.082$
         & $0.700\pm0.091$ & $0.668\pm0.088$ & $0.670\pm0.088$ & 32 \\
$W_{down}$ & $0.677\pm0.094$ & $0.535\pm0.157$ & $0.547\pm0.157$
           & $0.733\pm0.099$ & $0.590\pm0.171$ & $0.603\pm0.172$
           & $0.786\pm0.102$ & $0.646\pm0.185$ & $0.658\pm0.185$ & 32 \\
$W_{gate}$ & $0.396\pm0.061$ & $0.374\pm0.058$ & $0.376\pm0.059$
           & $0.433\pm0.067$ & $0.413\pm0.063$ & $0.415\pm0.064$
           & $0.471\pm0.073$ & $0.453\pm0.069$ & $0.456\pm0.070$ & 32 \\
\bottomrule
\end{tabular}%
}
\end{table}

\begin{table}[H]
\caption{Per-operator activation-conditioned output error (relative Frobenius; lower is better) under TT (cores4) tensorization with $\alpha=1$.
Entries are reported as mean $\mu$ and standard deviation $\sigma$ over $N=32$ single-operator trials per matrix type.}
\label{tab:perop_outerr_tt4}
\centering
\footnotesize
\setlength{\tabcolsep}{3.2pt}
\renewcommand{\arraystretch}{1.12}
\resizebox{\textwidth}{!}{%
\begin{tabular}{lccc ccc ccc c}
\toprule
& \multicolumn{3}{c}{2$\times$ compression} & \multicolumn{3}{c}{4$\times$ compression} & \multicolumn{3}{c}{6$\times$ compression} & \\
\cmidrule(lr){2-4}\cmidrule(lr){5-7}\cmidrule(lr){8-10}
Matrix
& Vanilla ($\mu\!\pm\!\sigma$) & Std ($\mu\!\pm\!\sigma$) & MeanAbs ($\mu\!\pm\!\sigma$)
& Vanilla ($\mu\!\pm\!\sigma$) & Std ($\mu\!\pm\!\sigma$) & MeanAbs ($\mu\!\pm\!\sigma$)
& Vanilla ($\mu\!\pm\!\sigma$) & Std ($\mu\!\pm\!\sigma$) & MeanAbs ($\mu\!\pm\!\sigma$)
& $N$ \\
\midrule
$W_q$      & $0.371\pm0.020$ & $0.301\pm0.030$ & $0.293\pm0.032$
           & $0.572\pm0.023$ & $0.474\pm0.041$ & $0.457\pm0.042$
           & $0.671\pm0.024$ & $0.568\pm0.045$ & $0.547\pm0.047$ & 32 \\
$W_k$      & $0.450\pm0.024$ & $0.383\pm0.037$ & $0.366\pm0.039$
           & $0.676\pm0.020$ & $0.609\pm0.045$ & $0.584\pm0.047$
           & $0.769\pm0.016$ & $0.709\pm0.046$ & $0.685\pm0.047$ & 32 \\
$W_v$      & $0.643\pm0.075$ & $0.541\pm0.024$ & $0.545\pm0.025$
           & $0.817\pm0.066$ & $0.733\pm0.026$ & $0.735\pm0.027$
           & $0.876\pm0.056$ & $0.807\pm0.025$ & $0.808\pm0.025$ & 32 \\
$W_o$      & $0.601\pm0.065$ & $0.473\pm0.101$ & $0.484\pm0.109$
           & $0.765\pm0.054$ & $0.651\pm0.103$ & $0.656\pm0.119$
           & $0.827\pm0.047$ & $0.728\pm0.101$ & $0.728\pm0.118$ & 32 \\
$W_{up}$   & $0.587\pm0.037$ & $0.564\pm0.032$ & $0.567\pm0.035$
           & $0.774\pm0.028$ & $0.754\pm0.024$ & $0.753\pm0.026$
           & $0.843\pm0.021$ & $0.826\pm0.019$ & $0.823\pm0.021$ & 32 \\
$W_{down}$ & $0.594\pm0.046$ & $0.528\pm0.128$ & $0.542\pm0.130$
           & $0.773\pm0.042$ & $0.696\pm0.161$ & $0.705\pm0.163$
           & $0.839\pm0.037$ & $0.758\pm0.172$ & $0.765\pm0.175$ & 32 \\
$W_{gate}$ & $0.449\pm0.034$ & $0.429\pm0.035$ & $0.428\pm0.037$
           & $0.620\pm0.038$ & $0.595\pm0.043$ & $0.589\pm0.045$
           & $0.688\pm0.041$ & $0.663\pm0.047$ & $0.654\pm0.049$ & 32 \\
\bottomrule
\end{tabular}%
}
\end{table}

\FloatBarrier
\subsection{Llama topology and alpha ablations}
\label{app:llama_ablation}

\begin{table}[H]
\caption{TT vs.\ TTN on WikiText at matched compression (parameter budget).
Entries are \textbf{within-backend medians} over the same set of single-operator trials (all blocks and matrices),
reported \emph{separately} for each backend (TT core4, TTN depth4, TTN depth8). This table is not pooled across backends; the main text reports pooled results. AWT value is the lower median across AWT methods.}

\label{tab:tt_baseline_comparison}
\centering
\small
\setlength{\tabcolsep}{5pt}
\begin{tabular}{l l cc cc}
\toprule
Comp. & Variant & PPL (Van.) & $\Delta$ & PPL (AWT) & $\Delta$ \\
\midrule
2$\times$ & TT (core4)     & 8.80 & 0.00 & 8.78 & 0.00 \\
2$\times$ & TTN (depth4)   & 8.73 & -0.07 & 8.72 & -0.06 \\
2$\times$ & TTN (depth8)   & 8.74 & -0.06 & 8.72 & -0.06 \\
\midrule
4$\times$ & TT (core4)     & 9.01 & 0.00 & 8.98 & 0.00 \\
4$\times$ & TTN (depth4)   & 8.82 & -0.19 & 8.79 & -0.19 \\
4$\times$ & TTN (depth8)   & 8.83 & -0.18 & 8.79 & -0.19 \\
\midrule
6$\times$ & TT (core4)     & 9.10 & 0.00 & 9.07 & 0.00 \\
6$\times$ & TTN (depth4)   & 8.87 & -0.23 & 8.84 & -0.23 \\
6$\times$ & TTN (depth8)   & 8.89 & -0.21 & 8.84 & -0.23 \\
\bottomrule
\end{tabular}
\end{table}

\begin{table}[H]
\caption{Effect of the reweighting exponent $\alpha$ at $2\times$ compression.
  AWT consistently reduces activation-conditioned output error (our functional fidelity proxy),
  while weight-space error exhibits a different optimum (continuing to improve at larger $\alpha$).}
\label{tab:alpha_ablation_2x}
\centering
\small
\setlength{\tabcolsep}{4pt}
\renewcommand{\arraystretch}{1.05}
\begin{tabular}{llrrrr}
\toprule
& & \multicolumn{2}{c}{Output error (median)} & \multicolumn{2}{c}{Weight error (median)} \\
\cmidrule(lr){3-4}\cmidrule(lr){5-6}
$\alpha$ & Method & Value & $\Delta$ vs Van. & Value & $\Delta$ vs Van. \\
\midrule
0.5 & Std     & 0.4471 & -15.7\% & 0.5985 & -1.0\%  \\
0.5 & MeanAbs & 0.4430 & -16.4\% & 0.5953 & -1.6\%  \\
1.0 & Std     & 0.4298 & -18.9\% & 0.5500 & -9.1\%  \\
1.0 & MeanAbs & 0.4339 & -18.2\% & 0.5309 & -12.2\% \\
1.5 & Std     & 0.4385 & -17.3\% & 0.3935 & -34.9\% \\
1.5 & MeanAbs & 0.4466 & -15.8\% & 0.3499 & -42.1\% \\
\bottomrule
\end{tabular}
\end{table}

\begin{table}[H]
\caption{Effect of the reweighting exponent $\alpha$ at $6\times$ compression.
AWT consistently reduces activation-conditioned output error (our functional fidelity proxy),
while weight-space error exhibits a different optimum (continuing to improve at larger $\alpha$).}
\label{tab:alpha_ablation_6x}
\centering
\small
\setlength{\tabcolsep}{4pt}
\renewcommand{\arraystretch}{1.05}
\begin{tabular}{llrrrr}
\toprule
& & \multicolumn{2}{c}{Output error (median)} & \multicolumn{2}{c}{Weight error (median)} \\
\cmidrule(lr){3-4}\cmidrule(lr){5-6}
$\alpha$ & Method & Value & $\Delta$ vs Van. & Value & $\Delta$ vs Van. \\
\midrule
0.5 & Std     & 0.6734 & -9.1\%  & 0.8212 & -1.4\%  \\
0.5 & MeanAbs & 0.6659 & -10.2\% & 0.8180 & -1.8\%  \\
1.0 & Std     & 0.6518 & -12.1\% & 0.7616 & -8.5\%  \\
1.0 & MeanAbs & 0.6477 & -12.6\% & 0.7374 & -11.4\% \\
1.5 & Std     & 0.6606 & -10.9\% & 0.5973 & -28.3\% \\
1.5 & MeanAbs & 0.6630 & -10.5\% & 0.5169 & -37.9\% \\
\bottomrule
\end{tabular}
\end{table}

\begin{table}[H]
\centering
\caption{Per-matrix Spearman correlation between $\Delta$RelFrob\_out and $\Delta$PPL
(AWT$-$Vanilla), $\alpha=1.0$.}
\label{tab:spearman_output_error_vs_ppl}
\begin{tabular}{lrr}
\toprule
Matrix & N & Spearman $\rho$ \\
\midrule
Wdown & 440 & 0.441 \\
Wgate & 448 & 0.026 \\
Wk & 570 & 0.490 \\
Wo & 576 & 0.284 \\
Wq & 569 & 0.702 \\
Wup & 448 & 0.383 \\
Wv & 576 & 0.281 \\
\bottomrule
\end{tabular}
\end{table}

\begin{table}[H]
\centering
\caption{Per-matrix Spearman correlation between $\Delta$Weight Error and $\Delta$PPL
(AWT$-$Vanilla), $\alpha=1.0$.}
\label{tab:spearman_weight_error_vs_ppl}
\begin{tabular}{lrr}
\toprule
Matrix & N & Spearman $\rho$ \\
\midrule
Wdown & 440 & 0.486 \\
Wgate & 448 & 0.129 \\
Wk & 570 & -0.178 \\
Wo & 576 & 0.034 \\
Wq & 569 & -0.395 \\
Wup & 448 & 0.163 \\
Wv & 576 & -0.056 \\
\bottomrule
\end{tabular}
\end{table}

\FloatBarrier
\subsection{Downstream evaluation on Llama~3.1~8B}
\label{app:downstream}

\begin{table}[H]
\caption{Downstream single-operator evaluation on Llama~3.1~8B. AWT improves
over vanilla tensorization at every tested compression on both HellaSwag and
ARC-Challenge. Higher is better.}
\label{tab:app_downstream_summary}
\centering
\small
\setlength{\tabcolsep}{4pt}
\begin{tabular}{llccccc}
\toprule
Task & Comp. & Dense & Vanilla & AWT & Method & Gap closed \\
\midrule
HellaSwag
& $2\times$ & 0.7931 & 0.7891 & 0.7894 & Std & 8.8\% \\
& $4\times$ & 0.7931 & 0.7867 & 0.7873 & Std & 9.4\% \\
& $6\times$ & 0.7931 & 0.7849 & 0.7857 & Std & 10.3\% \\
\midrule
ARC-Challenge
& $2\times$ & 0.5512 & 0.5486 & 0.5495 & Std & 33.3\% \\
& $4\times$ & 0.5512 & 0.5461 & 0.5469 & Std & 16.7\% \\
& $6\times$ & 0.5512 & 0.5435 & 0.5452 & Std & 22.2\% \\
\bottomrule
\end{tabular}
\end{table}

\begin{table}[H]
\caption{Backend-specific downstream results on Llama~3.1~8B under
single-operator replacement. Entries are within-backend pooled medians over the
canonical single-operator benchmark rows. $\Delta$ is measured against TT core4
at the same compression when available. Higher is better.}
\label{tab:app_downstream_backend}
\centering
\small
\setlength{\tabcolsep}{3.5pt}
\begin{tabular}{lllrrrr}
\toprule
Task & Comp. & Backend & Vanilla & $\Delta$ & AWT & $\Delta$ \\
\midrule
HellaSwag & $2\times$ & TTN depth4 & 0.7895 & +0.0007 & 0.7901 & +0.0013 \\
HellaSwag & $2\times$ & TTN depth8 & 0.7895 & +0.0007 & 0.7900 & +0.0012 \\
HellaSwag & $4\times$ & TT core4   & 0.7839 & +0.0000 & 0.7850 & +0.0000 \\
HellaSwag & $4\times$ & TTN depth4 & 0.7876 & +0.0038 & 0.7882 & +0.0032 \\
HellaSwag & $4\times$ & TTN depth8 & 0.7875 & +0.0037 & 0.7882 & +0.0032 \\
HellaSwag & $6\times$ & TT core4   & 0.7817 & +0.0000 & 0.7826 & +0.0000 \\
HellaSwag & $6\times$ & TTN depth4 & 0.7861 & +0.0044 & 0.7872 & +0.0046 \\
HellaSwag & $6\times$ & TTN depth8 & 0.7859 & +0.0042 & 0.7869 & +0.0043 \\
\midrule
ARC-Challenge & $2\times$ & TTN depth4 & 0.5495 & +0.0017 & 0.5495 & +0.0009 \\
ARC-Challenge & $2\times$ & TTN depth8 & 0.5495 & +0.0017 & 0.5495 & +0.0009 \\
ARC-Challenge & $4\times$ & TT core4   & 0.5418 & +0.0000 & 0.5435 & +0.0000 \\
ARC-Challenge & $4\times$ & TTN depth4 & 0.5478 & +0.0060 & 0.5486 & +0.0051 \\
ARC-Challenge & $4\times$ & TTN depth8 & 0.5478 & +0.0060 & 0.5478 & +0.0043 \\
ARC-Challenge & $6\times$ & TT core4   & 0.5384 & +0.0000 & 0.5414 & +0.0000 \\
ARC-Challenge & $6\times$ & TTN depth4 & 0.5461 & +0.0077 & 0.5469 & +0.0055 \\
ARC-Challenge & $6\times$ & TTN depth8 & 0.5461 & +0.0077 & 0.5469 & +0.0055 \\
\bottomrule
\end{tabular}
\end{table}

\FloatBarrier
\subsection{SVD and dense full-covariance comparisons}
\label{app:svd_cov}

\begin{table}[H]
\caption{Additional baselines and diagonal-design analysis. Left:
matched-compression SVD under the Llama~3.1~8B single-operator protocol.
Right: diagonal AWT versus dense full-covariance preconditioning.}
\label{tab:app_svd_cov}
\centering
\small
\begin{minipage}[t]{0.41\textwidth}
\centering
\textbf{SVD baseline}\\[0.35em]
\setlength{\tabcolsep}{4pt}
\begin{tabular}{lccc}
\toprule
Comp. & TT/TTN & AWT & SVD \\
\midrule
$2\times$ & 8.79 & 8.76 & 8.71 \\
$4\times$ & 8.90 & 8.85 & 8.81 \\
$6\times$ & 8.97 & 8.93 & 8.87 \\
\bottomrule
\end{tabular}
\end{minipage}
\hfill
\begin{minipage}[t]{0.55\textwidth}
\centering
\textbf{Diagonal vs. full covariance}\\[0.35em]
\setlength{\tabcolsep}{4pt}
\begin{tabular}{lc}
\toprule
Comparison & Result \\
\midrule
Diagonal covariance mass, L/M/Q & 0.032 / 0.142 / 0.212 \\
Off-diagonal Frobenius ratio, L/M/Q & 0.984 / 0.926 / 0.885 \\
Full-cov better on weighted objective & 80 / 81 \\
Diagonal better on held-out output error & 53 / 81 \\
\bottomrule
\end{tabular}
\end{minipage}
\end{table}

\begin{table}[H]
\caption{Diagonal AWT versus dense full-covariance preconditioning. The
activation covariance is strongly non-diagonal, and full covariance wins on its
own weighted objective, but diagonal AWT more often gives lower held-out local
output error.}
\label{tab:app_covariance}
\centering
\small
\setlength{\tabcolsep}{4pt}
\begin{tabular}{lc}
\toprule
Statistic / comparison & Result \\
\midrule
Diagonal covariance mass, Llama / Ministral / Qwen
& 0.0320 / 0.1415 / 0.2122 \\
Off-diagonal Frobenius ratio, Llama / Ministral / Qwen
& 0.9838 / 0.9255 / 0.8850 \\
Full-cov better on $\mathrm{RelFrob}_{\Sigma}$ objective
& 80 / 81 cases \\
Full-cov better on held-out $\mathrm{RelFrob}_{\mathrm{out}}$
& 28 / 81 cases \\
Diagonal better on held-out $\mathrm{RelFrob}_{\mathrm{out}}$
& 53 / 81 cases \\
\midrule
Llama: full-cov vs diagonal wins on held-out output error
& 13 vs 14 \\
Ministral: full-cov vs diagonal wins on held-out output error
& 10 vs 17 \\
Qwen: full-cov vs diagonal wins on held-out output error
& 5 vs 22 \\
Mean $\Delta$ output error, Llama / Ministral / Qwen
& -0.0284 / +0.6068 / +0.9093 \\
\bottomrule
\end{tabular}

\vspace{0.4em}
\footnotesize
$\Delta=\mathrm{RelFrob}_{\mathrm{out}}(\text{full-cov})
-\mathrm{RelFrob}_{\mathrm{out}}(\text{diagonal})$, so positive values favor
the diagonal method.
\end{table}

\FloatBarrier
\subsection{Cross-model single-operator details}
\label{app:cross_model_details}

\begin{table}[H]
\caption{Single-operator WikiText perplexity summaries for Ministral~8B and
Qwen2.5~7B. These are the per-family counterparts to the pooled cross-model
table in the main text. Lower PPL is better.}
\label{tab:app_cross_model_summaries}
\centering
\small
\setlength{\tabcolsep}{4pt}
\begin{tabular}{llccccc}
\toprule
Model & Comp. & Dense & Vanilla & AWT & Method & Gap closed \\
\midrule
Ministral~8B
& $2\times$ & 8.66 & 8.74 & 8.71 & Std & 35.3\% \\
& $4\times$ & 8.66 & 8.81 & 8.76 & Std & 31.3\% \\
& $6\times$ & 8.66 & 8.86 & 8.80 & Std & 28.6\% \\
\midrule
Qwen2.5~7B
& $2\times$ & 9.59 & 9.76 & 9.70 & Std & 33.5\% \\
& $4\times$ & 9.59 & 9.85 & 9.79 & Std & 25.8\% \\
& $6\times$ & 9.59 & 9.91 & 9.84 & Std & 22.2\% \\
\bottomrule
\end{tabular}
\end{table}

\begin{table}[H]
\caption{Spearman rank correlation between changes in layer-wise metrics and
$\Delta$PPL on WikiText for Ministral~8B and Qwen2.5~7B
($\alpha=1.0$). Groups: QK$=\{W_q,W_k\}$, VO$=\{W_v,W_o\}$,
MLP$=\{W_{up},W_{down},W_{gate}\}$.}
\label{tab:app_cross_model_spearman}
\centering
\small
\setlength{\tabcolsep}{4pt}
\begin{tabular}{llrrrr}
\toprule
Model & Metric & Overall & QK & VO & MLP \\
\midrule
Ministral~8B
& $\Delta\mathrm{RelFrob}_{\mathrm{out}}$ & 0.395 & 0.794 & 0.245 & 0.235 \\
& $\Delta\mathrm{RelFrob}_{\mathrm{w}}$   & 0.042 & -0.299 & 0.438 & -0.072 \\
\midrule
Qwen2.5~7B
& $\Delta\mathrm{RelFrob}_{\mathrm{out}}$ & 0.327 & 0.401 & 0.458 & 0.165 \\
& $\Delta\mathrm{RelFrob}_{\mathrm{w}}$   & -0.049 & -0.213 & -0.260 & 0.164 \\
\bottomrule
\end{tabular}
\end{table}

\FloatBarrier
\subsection{Cross-model alpha ablations}
\label{app:alpha_cross_model}

\begin{table}[H]
\caption{Alpha ablations for Ministral~8B. Entries report pooled medians over
single-operator sweep rows. Lower error is better.}
\label{tab:app_ministral_alpha}
\centering
\small
\setlength{\tabcolsep}{3pt}
\begin{tabular}{llrrrrrrrrrrrr}
\toprule
& & \multicolumn{4}{c}{$2\times$} & \multicolumn{4}{c}{$4\times$} & \multicolumn{4}{c}{$6\times$} \\
\cmidrule(lr){3-6}\cmidrule(lr){7-10}\cmidrule(lr){11-14}
$\alpha$ & Method
& Out & $\Delta$ & Weight & $\Delta$
& Out & $\Delta$ & Weight & $\Delta$
& Out & $\Delta$ & Weight & $\Delta$ \\
\midrule
0.5 & Std
& 0.398 & -15.1 & 0.585 & +0.1
& 0.518 & -14.7 & 0.736 & -2.7
& 0.584 & -12.2 & 0.805 & -2.4 \\
0.5 & MeanAbs
& 0.395 & -15.8 & 0.582 & -0.4
& 0.517 & -14.9 & 0.734 & -3.0
& 0.582 & -12.5 & 0.802 & -2.7 \\
1.0 & Std
& 0.376 & -19.9 & 0.525 & -10.2
& 0.500 & -17.7 & 0.660 & -12.8
& 0.562 & -15.5 & 0.726 & -12.0 \\
1.0 & MeanAbs
& 0.377 & -19.7 & 0.502 & -14.1
& 0.500 & -17.8 & 0.635 & -16.1
& 0.564 & -15.2 & 0.699 & -15.2 \\
1.5 & Std
& 0.380 & -18.9 & 0.365 & -37.5
& 0.505 & -16.9 & 0.477 & -36.9
& 0.570 & -14.2 & 0.549 & -33.5 \\
1.5 & MeanAbs
& 0.387 & -17.5 & 0.285 & -51.3
& 0.509 & -16.2 & 0.379 & -49.9
& 0.574 & -13.6 & 0.430 & -47.8 \\
\bottomrule
\end{tabular}

\vspace{0.4em}
\footnotesize
Vanilla pooled medians: output / weight error =
$0.469/0.585$ at $2\times$, $0.608/0.756$ at $4\times$,
and $0.665/0.824$ at $6\times$. $\Delta$ values are percentages.
\end{table}

\begin{table}[H]
\caption{Alpha ablations for Qwen2.5~7B. Entries report pooled medians over
single-operator sweep rows. Lower error is better.}
\label{tab:app_qwen_alpha}
\centering
\small
\setlength{\tabcolsep}{3pt}
\begin{tabular}{llrrrrrrrrrrrr}
\toprule
& & \multicolumn{4}{c}{$2\times$} & \multicolumn{4}{c}{$4\times$} & \multicolumn{4}{c}{$6\times$} \\
\cmidrule(lr){3-6}\cmidrule(lr){7-10}\cmidrule(lr){11-14}
$\alpha$ & Method
& Out & $\Delta$ & Weight & $\Delta$
& Out & $\Delta$ & Weight & $\Delta$
& Out & $\Delta$ & Weight & $\Delta$ \\
\midrule
0.5 & Std
& 0.377 & -18.6 & 0.594 & +0.8
& 0.522 & -16.8 & 0.745 & -2.6
& 0.590 & -15.1 & 0.814 & -2.2 \\
0.5 & MeanAbs
& 0.375 & -19.1 & 0.588 & -0.1
& 0.515 & -17.9 & 0.740 & -3.3
& 0.579 & -16.7 & 0.808 & -3.0 \\
1.0 & Std
& 0.366 & -21.1 & 0.487 & -17.3
& 0.501 & -20.0 & 0.627 & -18.1
& 0.568 & -18.4 & 0.702 & -15.7 \\
1.0 & MeanAbs
& 0.366 & -21.0 & 0.445 & -24.4
& 0.505 & -19.4 & 0.578 & -24.4
& 0.568 & -18.4 & 0.654 & -21.4 \\
1.5 & Std
& 0.395 & -14.8 & 0.271 & -54.0
& 0.520 & -16.9 & 0.364 & -52.5
& 0.582 & -16.3 & 0.409 & -50.9 \\
1.5 & MeanAbs
& 0.405 & -12.6 & 0.196 & -66.7
& 0.527 & -16.0 & 0.286 & -62.7
& 0.587 & -15.6 & 0.332 & -60.1 \\
\bottomrule
\end{tabular}

\vspace{0.4em}
\footnotesize
Vanilla pooled medians: output / weight error =
$0.463/0.589$ at $2\times$, $0.627/0.765$ at $4\times$,
and $0.695/0.833$ at $6\times$. $\Delta$ values are percentages.
\end{table}

\FloatBarrier
\subsection{Cross-model backend comparisons}
\label{app:backend_cross_model}

\begin{table}[H]
\caption{Backend comparison on WikiText under single-operator replacement for
Ministral~8B. Entries are within-backend pooled medians. $\Delta$ is measured
against TT core4 at the same compression. Lower PPL is better.}
\label{tab:app_ministral_backend}
\centering
\small
\setlength{\tabcolsep}{4pt}
\begin{tabular}{llrrrr}
\toprule
Comp. & Backend & PPL (Van.) & $\Delta$ & PPL (AWT) & $\Delta$ \\
\midrule
$2\times$ & TT core4   & 8.75 & +0.00 & 8.71 & +0.00 \\
$2\times$ & TTN depth4 & 8.72 & -0.03 & 8.70 & -0.01 \\
$2\times$ & TTN depth8 & 8.72 & -0.03 & 8.70 & -0.01 \\
$4\times$ & TT core4   & 8.86 & +0.00 & 8.79 & +0.00 \\
$4\times$ & TTN depth4 & 8.77 & -0.09 & 8.75 & -0.04 \\
$4\times$ & TTN depth8 & 8.78 & -0.08 & 8.75 & -0.04 \\
$6\times$ & TT core4   & 8.92 & +0.00 & 8.84 & +0.00 \\
$6\times$ & TTN depth4 & 8.81 & -0.11 & 8.78 & -0.06 \\
$6\times$ & TTN depth8 & 8.81 & -0.10 & 8.78 & -0.06 \\
\bottomrule
\end{tabular}
\end{table}

\begin{table}[H]
\caption{Backend comparison on WikiText under single-operator replacement for
Qwen2.5~7B. Entries are within-backend pooled medians. $\Delta$ is measured
against TT core4 at the same compression. Lower PPL is better.}
\label{tab:app_qwen_backend}
\centering
\small
\setlength{\tabcolsep}{4pt}
\begin{tabular}{llrrrr}
\toprule
Comp. & Backend & PPL (Van.) & $\Delta$ & PPL (AWT) & $\Delta$ \\
\midrule
$2\times$ & TT core4   & 9.75 & +0.00 & 9.72 & +0.00 \\
$2\times$ & TTN depth4 & 9.76 & +0.01 & 9.69 & -0.02 \\
$2\times$ & TTN depth8 & 9.78 & +0.02 & 9.70 & -0.02 \\
$4\times$ & TT core4   & 9.88 & +0.00 & 9.82 & +0.00 \\
$4\times$ & TTN depth4 & 9.84 & -0.04 & 9.75 & -0.07 \\
$4\times$ & TTN depth8 & 9.85 & -0.03 & 9.77 & -0.05 \\
$6\times$ & TT core4   & 9.95 & +0.00 & 9.89 & +0.00 \\
$6\times$ & TTN depth4 & 9.88 & -0.07 & 9.81 & -0.08 \\
$6\times$ & TTN depth8 & 9.91 & -0.04 & 9.81 & -0.08 \\
\bottomrule
\end{tabular}
\end{table}

\FloatBarrier
\subsection{Multi-operator replacement details}
\label{app:multiop_details}

\begin{table}[H]
\caption{Multi-operator WikiText suffix replacement on Llama~3.1~8B. The
attention group replaces $\{W_q,W_k,W_v,W_o\}$ jointly in the last $k$ blocks.
Lower PPL is better; negative $\Delta$PPL favors AWT.}
\label{tab:app_multiop_wikitext_attn}
\centering
\small
\setlength{\tabcolsep}{4pt}
\begin{tabular}{llrrrrl}
\toprule
$k$ & Comp. & Vanilla PPL & AWT PPL & $\Delta$PPL & Gap closed & Finite backends \\
\midrule
1 & $2\times$ & 9.33 & 8.92 & -0.41 & 59.5\% & TT core4, TTN depth4, TTN depth8 \\
1 & $4\times$ & 9.58 & 9.12 & -0.47 & 49.0\% & TT core4, TTN depth4, TTN depth8 \\
1 & $6\times$ & 9.71 & 9.33 & -0.46 & 35.3\% & TT core4, TTN depth4, TTN depth8 \\
2 & $2\times$ & 9.78 & 9.19 & -0.59 & 51.8\% & TT core4, TTN depth4, TTN depth8 \\
2 & $4\times$ & 10.28 & 9.60 & -0.72 & 41.3\% & TT core4, TTN depth4, TTN depth8 \\
2 & $6\times$ & 10.56 & 9.91 & -0.81 & 33.9\% & TT core4, TTN depth4, TTN depth8 \\
4 & $2\times$ & 10.56 & 9.80 & -0.76 & 39.4\% & TT core4, TTN depth4, TTN depth8 \\
4 & $4\times$ & 11.93 & 10.92 & -1.02 & 30.8\% & TT core4, TTN depth4, TTN depth8 \\
4 & $6\times$ & 12.68 & 11.57 & -1.11 & 27.4\% & TT core4, TTN depth4, TTN depth8 \\
8 & $2\times$ & 11.85 & 10.45 & -1.40 & 43.6\% & TT core4, TTN depth4, TTN depth8 \\
8 & $4\times$ & 15.55 & 13.08 & -2.48 & 35.8\% & TT core4, TTN depth4, TTN depth8 \\
8 & $6\times$ & 17.55 & 14.81 & -2.75 & 30.8\% & TT core4, TTN depth4, TTN depth8 \\
\bottomrule
\end{tabular}
\end{table}

\begin{table}[H]
\caption{Multi-operator WikiText suffix replacement on Llama~3.1~8B when all
seven dense matrices are replaced jointly in the last $k$ blocks. Lower PPL is
better; negative $\Delta$PPL favors AWT.}
\label{tab:app_multiop_wikitext_all}
\centering
\small
\setlength{\tabcolsep}{4pt}
\begin{tabular}{llrrrr}
\toprule
$k$ & Comp. & Vanilla PPL & AWT PPL & $\Delta$PPL & Gap closed \\
\midrule
1 & $2\times$ & 13.87 & 11.99 & -1.86 & 36.0\% \\
1 & $4\times$ & 15.54 & 12.76 & -2.61 & 40.2\% \\
1 & $6\times$ & 17.65 & 13.60 & -3.62 & 45.0\% \\
2 & $2\times$ & 21.53 & 13.79 & -7.74 & 60.0\% \\
2 & $4\times$ & 24.47 & 15.09 & -9.21 & 59.3\% \\
2 & $6\times$ & 30.18 & 17.28 & -11.71 & 59.9\% \\
\bottomrule
\end{tabular}
\end{table}

\begin{table}[H]
\caption{Multi-operator WikiText suffix replacement on Llama~3.1~8B when the
MLP group $\{W_{up},W_{gate},W_{down}\}$ is replaced jointly. Empty rows in the
source runs correspond to non-finite PPL. Lower PPL is better; negative
$\Delta$PPL favors AWT.}
\label{tab:app_multiop_wikitext_mlp}
\centering
\small
\setlength{\tabcolsep}{4pt}
\begin{tabular}{llrrrrl}
\toprule
$k$ & Comp. & Vanilla PPL & AWT PPL & $\Delta$PPL & Gap closed & Finite backends \\
\midrule
1 & $2\times$ & 13.54 & 12.27 & -1.27 & 26.0\% & TT core4, TTN depth4, TTN depth8 \\
1 & $4\times$ & 14.19 & 12.73 & -1.45 & 26.4\% & TT core4, TTN depth4, TTN depth8 \\
1 & $6\times$ & 14.78 & 13.25 & -1.53 & 24.9\% & TT core4, TTN depth4, TTN depth8 \\
2 & $2\times$ & 17.65 & 14.21 & -3.43 & 38.1\% & TT core4, TTN depth4, TTN depth8 \\
2 & $4\times$ & 18.52 & 14.67 & -3.79 & 39.0\% & TT core4, TTN depth4, TTN depth8 \\
2 & $6\times$ & 19.78 & 15.30 & -4.47 & 40.2\% & TT core4, TTN depth4, TTN depth8 \\
4 & $2\times$ & 30.38 & 17.41 & -12.97 & 59.6\% & TT core4, TTN depth4, TTN depth8 \\
4 & $4\times$ & 34.02 & 18.33 & -15.69 & 61.8\% & TTN depth4, TTN depth8 \\
4 & $6\times$ & 39.31 & 19.61 & -19.70 & 64.2\% & TTN depth4, TTN depth8 \\
8 & $2\times$ & 36.21 & 38.31 & +2.09 & -7.6\% & TT core4 \\
\bottomrule
\end{tabular}
\end{table}

\FloatBarrier
\subsection{Multi-operator BoolQ details}
\label{app:multiop_boolq}

\begin{table}[H]
\caption{Multi-operator BoolQ suffix replacement on Llama~3.1~8B. Attention
group replaces $\{W_q,W_k,W_v,W_o\}$ jointly in the last $k$ blocks. Higher is
better; positive $\Delta$ favors AWT.}
\label{tab:app_multiop_boolq_attn}
\centering
\small
\setlength{\tabcolsep}{4pt}
\begin{tabular}{llrrrrl}
\toprule
$k$ & Comp. & Vanilla & AWT & $\Delta$ & Error reduction vs vanilla & Finite backends \\
\midrule
1 & $2\times$ & 0.7900 & 0.8000 & +0.0100 & 4.8\% & TT core4, TTN depth4, TTN depth8 \\
1 & $4\times$ & 0.8000 & 0.7900 & -0.0100 & -5.0\% & TT core4, TTN depth4, TTN depth8 \\
1 & $6\times$ & 0.8000 & 0.8000 & +0.0000 & 0.0\% & TT core4, TTN depth4, TTN depth8 \\
2 & $2\times$ & 0.8000 & 0.8100 & +0.0100 & 5.0\% & TT core4, TTN depth4, TTN depth8 \\
2 & $4\times$ & 0.8100 & 0.8300 & +0.0200 & 10.5\% & TT core4, TTN depth4, TTN depth8 \\
2 & $6\times$ & 0.8100 & 0.8100 & +0.0000 & 0.0\% & TT core4, TTN depth4, TTN depth8 \\
4 & $2\times$ & 0.8000 & 0.8300 & +0.0200 & 15.0\% & TT core4, TTN depth4, TTN depth8 \\
4 & $4\times$ & 0.8200 & 0.8300 & +0.0100 & 5.6\% & TT core4, TTN depth4, TTN depth8 \\
4 & $6\times$ & 0.8100 & 0.8300 & +0.0200 & 10.5\% & TT core4, TTN depth4, TTN depth8 \\
8 & $2\times$ & 0.8100 & 0.8300 & +0.0200 & 10.5\% & TT core4, TTN depth4, TTN depth8 \\
8 & $4\times$ & 0.8200 & 0.8300 & +0.0100 & 5.6\% & TT core4, TTN depth4, TTN depth8 \\
8 & $6\times$ & 0.8300 & 0.8400 & +0.0300 & 5.9\% & TT core4, TTN depth4, TTN depth8 \\
\bottomrule
\end{tabular}
\end{table}

\begin{table}[H]
\caption{Multi-operator BoolQ suffix replacement on Llama~3.1~8B when all seven
dense matrices are replaced jointly in the last $k$ blocks. Higher is better.}
\label{tab:app_multiop_boolq_all}
\centering
\small
\setlength{\tabcolsep}{4pt}
\begin{tabular}{llrrrr}
\toprule
$k$ & Comp. & Vanilla & AWT & $\Delta$ & Error reduction vs vanilla \\
\midrule
1 & $2\times$ & 0.8100 & 0.8300 & +0.0200 & 10.5\% \\
1 & $4\times$ & 0.7900 & 0.8200 & +0.0300 & 14.3\% \\
1 & $6\times$ & 0.7600 & 0.8200 & +0.0600 & 25.0\% \\
2 & $2\times$ & 0.7900 & 0.7800 & -0.0100 & -4.8\% \\
2 & $4\times$ & 0.7800 & 0.7900 & +0.0100 & 4.5\% \\
2 & $6\times$ & 0.7900 & 0.7800 & +0.0000 & -4.8\% \\
\bottomrule
\end{tabular}
\end{table}

\begin{table}[H]
\caption{Multi-operator BoolQ suffix replacement on Llama~3.1~8B when the MLP
group $\{W_{up},W_{gate},W_{down}\}$ is replaced jointly. Higher is better.}
\label{tab:app_multiop_boolq_mlp}
\centering
\small
\setlength{\tabcolsep}{4pt}
\begin{tabular}{llrrrrl}
\toprule
$k$ & Comp. & Vanilla & AWT & $\Delta$ & Error reduction vs vanilla & Finite backends \\
\midrule
1 & $2\times$ & 0.8200 & 0.8200 & +0.0000 & 0.0\% & TT core4, TTN depth4, TTN depth8 \\
1 & $4\times$ & 0.8100 & 0.8200 & +0.0100 & 5.3\% & TT core4, TTN depth4, TTN depth8 \\
1 & $6\times$ & 0.8100 & 0.8300 & +0.0200 & 10.5\% & TT core4, TTN depth4, TTN depth8 \\
2 & $2\times$ & 0.7800 & 0.7800 & +0.0000 & 0.0\% & TT core4, TTN depth4, TTN depth8 \\
2 & $4\times$ & 0.7800 & 0.7800 & +0.0000 & 0.0\% & TT core4, TTN depth4, TTN depth8 \\
2 & $6\times$ & 0.7900 & 0.7800 & -0.0100 & -4.8\% & TT core4, TTN depth4, TTN depth8 \\
4 & $2\times$ & 0.7900 & 0.7900 & +0.0000 & 0.0\% & TT core4, TTN depth4, TTN depth8 \\
4 & $4\times$ & 0.7700 & 0.7800 & +0.0100 & 4.3\% & TT core4, TTN depth4, TTN depth8 \\
4 & $6\times$ & 0.7600 & 0.7900 & +0.0300 & 12.5\% & TT core4, TTN depth4, TTN depth8 \\
8 & $2\times$ & 0.7000 & 0.7100 & +0.0100 & 3.3\% & TT core4, TTN depth4, TTN depth8 \\
8 & $4\times$ & 0.7000 & 0.7000 & +0.0000 & 0.0\% & TT core4, TTN depth4, TTN depth8 \\
8 & $6\times$ & 0.7000 & 0.7200 & +0.0200 & 6.7\% & TT core4, TTN depth4, TTN depth8 \\
\bottomrule
\end{tabular}
\end{table}

\FloatBarrier
\subsection{Per-operator output-error breakdowns for Ministral and Qwen}
\label{app:per_operator_cross_model}

\begin{table}[H]
\caption{Per-operator activation-conditioned output error for Ministral~8B
under TTN depth4 tensorization with $\alpha=1$. Entries report mean $\mu$ and
standard deviation $\sigma$ over per-block sweep values. Lower is better.}
\label{tab:app_ministral_ttn4_output}
\centering
\footnotesize
\setlength{\tabcolsep}{3.2pt}
\renewcommand{\arraystretch}{1.12}
\resizebox{\textwidth}{!}{%
\begin{tabular}{lccc ccc ccc c}
\toprule
& \multicolumn{3}{c}{$2\times$ compression}
& \multicolumn{3}{c}{$4\times$ compression}
& \multicolumn{3}{c}{$6\times$ compression}
& \\
\cmidrule(lr){2-4}\cmidrule(lr){5-7}\cmidrule(lr){8-10}
Matrix
& Vanilla ($\mu\!\pm\!\sigma$) & Std ($\mu\!\pm\!\sigma$) & MeanAbs ($\mu\!\pm\!\sigma$)
& Vanilla ($\mu\!\pm\!\sigma$) & Std ($\mu\!\pm\!\sigma$) & MeanAbs ($\mu\!\pm\!\sigma$)
& Vanilla ($\mu\!\pm\!\sigma$) & Std ($\mu\!\pm\!\sigma$) & MeanAbs ($\mu\!\pm\!\sigma$)
& $N$ \\
\midrule
$W_q$
& $0.192\pm0.044$ & $0.158\pm0.038$ & $0.159\pm0.038$
& $0.258\pm0.052$ & $0.217\pm0.046$ & $0.218\pm0.046$
& $0.291\pm0.054$ & $0.248\pm0.049$ & $0.249\pm0.049$ & 36 \\
$W_k$
& $0.191\pm0.037$ & $0.151\pm0.032$ & $0.152\pm0.032$
& $0.289\pm0.044$ & $0.231\pm0.039$ & $0.233\pm0.039$
& $0.339\pm0.044$ & $0.272\pm0.041$ & $0.275\pm0.041$ & 36 \\
$W_v$
& $0.558\pm0.080$ & $0.458\pm0.061$ & $0.458\pm0.060$
& $0.708\pm0.081$ & $0.608\pm0.068$ & $0.608\pm0.068$
& $0.770\pm0.079$ & $0.673\pm0.067$ & $0.675\pm0.067$ & 36 \\
$W_o$
& $0.492\pm0.111$ & $0.375\pm0.099$ & $0.380\pm0.097$
& $0.625\pm0.135$ & $0.506\pm0.124$ & $0.508\pm0.122$
& $0.680\pm0.144$ & $0.569\pm0.135$ & $0.571\pm0.133$ & 36 \\
$W_{up}$
& $0.559\pm0.078$ & $0.520\pm0.079$ & $0.522\pm0.080$
& $0.617\pm0.087$ & $0.577\pm0.088$ & $0.579\pm0.088$
& $0.669\pm0.094$ & $0.628\pm0.095$ & $0.631\pm0.095$ & 36 \\
$W_{down}$
& $0.655\pm0.090$ & $0.507\pm0.149$ & $0.523\pm0.153$
& $0.716\pm0.096$ & $0.568\pm0.166$ & $0.585\pm0.171$
& $0.769\pm0.101$ & $0.625\pm0.180$ & $0.642\pm0.186$ & 36 \\
$W_{gate}$
& $0.348\pm0.050$ & $0.328\pm0.050$ & $0.329\pm0.050$
& $0.387\pm0.055$ & $0.366\pm0.056$ & $0.368\pm0.056$
& $0.423\pm0.061$ & $0.403\pm0.061$ & $0.404\pm0.061$ & 36 \\
\bottomrule
\end{tabular}%
}
\end{table}

\begin{table}[H]
\caption{Per-operator activation-conditioned output error for Ministral~8B
under TTN depth8 tensorization with $\alpha=1$. Entries report mean $\mu$ and
standard deviation $\sigma$ over per-block sweep values. Lower is better.}
\label{tab:app_ministral_ttn8_output}
\centering
\footnotesize
\setlength{\tabcolsep}{3.2pt}
\renewcommand{\arraystretch}{1.12}
\resizebox{\textwidth}{!}{%
\begin{tabular}{lccc ccc ccc c}
\toprule
& \multicolumn{3}{c}{$2\times$ compression}
& \multicolumn{3}{c}{$4\times$ compression}
& \multicolumn{3}{c}{$6\times$ compression}
& \\
\cmidrule(lr){2-4}\cmidrule(lr){5-7}\cmidrule(lr){8-10}
Matrix
& Vanilla ($\mu\!\pm\!\sigma$) & Std ($\mu\!\pm\!\sigma$) & MeanAbs ($\mu\!\pm\!\sigma$)
& Vanilla ($\mu\!\pm\!\sigma$) & Std ($\mu\!\pm\!\sigma$) & MeanAbs ($\mu\!\pm\!\sigma$)
& Vanilla ($\mu\!\pm\!\sigma$) & Std ($\mu\!\pm\!\sigma$) & MeanAbs ($\mu\!\pm\!\sigma$)
& $N$ \\
\midrule
$W_q$
& $0.193\pm0.044$ & $0.160\pm0.038$ & $0.160\pm0.038$
& $0.260\pm0.052$ & $0.220\pm0.046$ & $0.220\pm0.046$
& $0.294\pm0.054$ & $0.251\pm0.049$ & $0.252\pm0.049$ & 36 \\
$W_k$
& $0.195\pm0.038$ & $0.155\pm0.032$ & $0.156\pm0.032$
& $0.293\pm0.044$ & $0.235\pm0.038$ & $0.236\pm0.039$
& $0.338\pm0.043$ & $0.274\pm0.039$ & $0.276\pm0.040$ & 36 \\
$W_v$
& $0.566\pm0.080$ & $0.466\pm0.061$ & $0.465\pm0.061$
& $0.718\pm0.079$ & $0.612\pm0.066$ & $0.613\pm0.066$
& $0.778\pm0.077$ & $0.672\pm0.066$ & $0.673\pm0.066$ & 36 \\
$W_o$
& $0.495\pm0.112$ & $0.378\pm0.100$ & $0.383\pm0.097$
& $0.629\pm0.136$ & $0.510\pm0.125$ & $0.513\pm0.123$
& $0.685\pm0.145$ & $0.575\pm0.136$ & $0.577\pm0.135$ & 36 \\
$W_{up}$
& $0.559\pm0.078$ & $0.521\pm0.079$ & $0.522\pm0.080$
& $0.618\pm0.087$ & $0.578\pm0.088$ & $0.580\pm0.088$
& $0.670\pm0.094$ & $0.630\pm0.095$ & $0.632\pm0.095$ & 36 \\
$W_{down}$
& $0.655\pm0.090$ & $0.507\pm0.149$ & $0.523\pm0.153$
& $0.717\pm0.096$ & $0.569\pm0.166$ & $0.586\pm0.171$
& $0.770\pm0.101$ & $0.626\pm0.180$ & $0.644\pm0.186$ & 36 \\
$W_{gate}$
& $0.348\pm0.050$ & $0.328\pm0.050$ & $0.329\pm0.050$
& $0.388\pm0.055$ & $0.367\pm0.056$ & $0.368\pm0.056$
& $0.424\pm0.061$ & $0.404\pm0.061$ & $0.405\pm0.061$ & 36 \\
\bottomrule
\end{tabular}%
}
\end{table}

\begin{table}[H]
\caption{Per-operator activation-conditioned output error for Ministral~8B
under TT core4 tensorization with $\alpha=1$. Entries report mean $\mu$ and
standard deviation $\sigma$ over per-block sweep values. Lower is better.}
\label{tab:app_ministral_ttcore4_output}
\centering
\footnotesize
\setlength{\tabcolsep}{3.2pt}
\renewcommand{\arraystretch}{1.12}
\resizebox{\textwidth}{!}{%
\begin{tabular}{lccc ccc ccc c}
\toprule
& \multicolumn{3}{c}{$2\times$ compression}
& \multicolumn{3}{c}{$4\times$ compression}
& \multicolumn{3}{c}{$6\times$ compression}
& \\
\cmidrule(lr){2-4}\cmidrule(lr){5-7}\cmidrule(lr){8-10}
Matrix
& Vanilla ($\mu\!\pm\!\sigma$) & Std ($\mu\!\pm\!\sigma$) & MeanAbs ($\mu\!\pm\!\sigma$)
& Vanilla ($\mu\!\pm\!\sigma$) & Std ($\mu\!\pm\!\sigma$) & MeanAbs ($\mu\!\pm\!\sigma$)
& Vanilla ($\mu\!\pm\!\sigma$) & Std ($\mu\!\pm\!\sigma$) & MeanAbs ($\mu\!\pm\!\sigma$)
& $N$ \\
\midrule
$W_q$
& $0.333\pm0.019$ & $0.228\pm0.024$ & $0.227\pm0.024$
& $0.489\pm0.024$ & $0.331\pm0.028$ & $0.331\pm0.029$
& $0.565\pm0.029$ & $0.382\pm0.032$ & $0.381\pm0.032$ & 36 \\
$W_k$
& $0.321\pm0.044$ & $0.219\pm0.022$ & $0.219\pm0.022$
& $0.500\pm0.065$ & $0.326\pm0.031$ & $0.325\pm0.031$
& $0.581\pm0.071$ & $0.378\pm0.037$ & $0.377\pm0.036$ & 36 \\
$W_v$
& $0.616\pm0.039$ & $0.448\pm0.043$ & $0.448\pm0.043$
& $0.780\pm0.046$ & $0.611\pm0.057$ & $0.611\pm0.057$
& $0.841\pm0.045$ & $0.677\pm0.061$ & $0.677\pm0.061$ & 36 \\
$W_o$
& $0.551\pm0.124$ & $0.449\pm0.119$ & $0.466\pm0.121$
& $0.698\pm0.149$ & $0.609\pm0.144$ & $0.620\pm0.148$
& $0.753\pm0.158$ & $0.677\pm0.153$ & $0.684\pm0.157$ & 36 \\
$W_{up}$
& $0.575\pm0.043$ & $0.493\pm0.061$ & $0.493\pm0.061$
& $0.748\pm0.038$ & $0.652\pm0.072$ & $0.651\pm0.072$
& $0.814\pm0.033$ & $0.713\pm0.074$ & $0.712\pm0.074$ & 36 \\
$W_{down}$
& $0.570\pm0.080$ & $0.470\pm0.139$ & $0.464\pm0.137$
& $0.722\pm0.124$ & $0.626\pm0.177$ & $0.623\pm0.175$
& $0.777\pm0.142$ & $0.688\pm0.189$ & $0.687\pm0.187$ & 36 \\
$W_{gate}$
& $0.396\pm0.040$ & $0.331\pm0.053$ & $0.331\pm0.053$
& $0.535\pm0.048$ & $0.446\pm0.071$ & $0.445\pm0.071$
& $0.592\pm0.051$ & $0.493\pm0.077$ & $0.492\pm0.077$ & 36 \\
\bottomrule
\end{tabular}%
}
\end{table}

\begin{table}[H]
\caption{Per-operator activation-conditioned output error for Qwen2.5~7B under
TTN depth4 tensorization with $\alpha=1$. Entries report mean $\mu$ and
standard deviation $\sigma$ over per-block sweep values. Lower is better.}
\label{tab:app_qwen_ttn4_output}
\centering
\footnotesize
\setlength{\tabcolsep}{3.2pt}
\renewcommand{\arraystretch}{1.12}
\resizebox{\textwidth}{!}{%
\begin{tabular}{lccc ccc ccc c}
\toprule
& \multicolumn{3}{c}{$2\times$ compression}
& \multicolumn{3}{c}{$4\times$ compression}
& \multicolumn{3}{c}{$6\times$ compression}
& \\
\cmidrule(lr){2-4}\cmidrule(lr){5-7}\cmidrule(lr){8-10}
Matrix
& Vanilla ($\mu\!\pm\!\sigma$) & Std ($\mu\!\pm\!\sigma$) & MeanAbs ($\mu\!\pm\!\sigma$)
& Vanilla ($\mu\!\pm\!\sigma$) & Std ($\mu\!\pm\!\sigma$) & MeanAbs ($\mu\!\pm\!\sigma$)
& Vanilla ($\mu\!\pm\!\sigma$) & Std ($\mu\!\pm\!\sigma$) & MeanAbs ($\mu\!\pm\!\sigma$)
& $N$ \\
\midrule
$W_q$
& $0.331\pm0.079$ & $0.222\pm0.069$ & $0.223\pm0.069$
& $0.446\pm0.099$ & $0.313\pm0.089$ & $0.313\pm0.089$
& $0.505\pm0.109$ & $0.363\pm0.099$ & $0.363\pm0.099$ & 28 \\
$W_k$
& $0.354\pm0.060$ & $0.207\pm0.048$ & $0.208\pm0.048$
& $0.566\pm0.076$ & $0.346\pm0.061$ & $0.347\pm0.060$
& $0.652\pm0.075$ & $0.419\pm0.063$ & $0.419\pm0.063$ & 28 \\
$W_v$
& $0.665\pm0.093$ & $0.469\pm0.119$ & $0.468\pm0.118$
& $0.823\pm0.072$ & $0.647\pm0.097$ & $0.644\pm0.102$
& $0.876\pm0.060$ & $0.728\pm0.082$ & $0.726\pm0.084$ & 28 \\
$W_o$
& $0.473\pm0.097$ & $0.375\pm0.080$ & $0.377\pm0.080$
& $0.619\pm0.121$ & $0.510\pm0.106$ & $0.510\pm0.106$
& $0.684\pm0.132$ & $0.574\pm0.119$ & $0.573\pm0.119$ & 28 \\
$W_{up}$
& $0.602\pm0.097$ & $0.469\pm0.119$ & $0.470\pm0.119$
& $0.659\pm0.098$ & $0.527\pm0.126$ & $0.528\pm0.126$
& $0.710\pm0.100$ & $0.580\pm0.132$ & $0.581\pm0.132$ & 28 \\
$W_{down}$
& $0.447\pm0.184$ & $0.284\pm0.184$ & $0.298\pm0.189$
& $0.490\pm0.200$ & $0.323\pm0.207$ & $0.337\pm0.212$
& $0.528\pm0.213$ & $0.360\pm0.228$ & $0.373\pm0.233$ & 28 \\
$W_{gate}$
& $0.312\pm0.130$ & $0.256\pm0.115$ & $0.257\pm0.116$
& $0.343\pm0.141$ & $0.287\pm0.127$ & $0.288\pm0.128$
& $0.371\pm0.150$ & $0.316\pm0.138$ & $0.317\pm0.138$ & 28 \\
\bottomrule
\end{tabular}%
}
\end{table}

\begin{table}[H]
\caption{Per-operator activation-conditioned output error for Qwen2.5~7B under
TTN depth8 tensorization with $\alpha=1$. Entries report mean $\mu$ and
standard deviation $\sigma$ over per-block sweep values. Lower is better.}
\label{tab:app_qwen_ttn8_output}
\centering
\footnotesize
\setlength{\tabcolsep}{3.2pt}
\renewcommand{\arraystretch}{1.12}
\resizebox{\textwidth}{!}{%
\begin{tabular}{lccc ccc ccc c}
\toprule
& \multicolumn{3}{c}{$2\times$ compression}
& \multicolumn{3}{c}{$4\times$ compression}
& \multicolumn{3}{c}{$6\times$ compression}
& \\
\cmidrule(lr){2-4}\cmidrule(lr){5-7}\cmidrule(lr){8-10}
Matrix
& Vanilla ($\mu\!\pm\!\sigma$) & Std ($\mu\!\pm\!\sigma$) & MeanAbs ($\mu\!\pm\!\sigma$)
& Vanilla ($\mu\!\pm\!\sigma$) & Std ($\mu\!\pm\!\sigma$) & MeanAbs ($\mu\!\pm\!\sigma$)
& Vanilla ($\mu\!\pm\!\sigma$) & Std ($\mu\!\pm\!\sigma$) & MeanAbs ($\mu\!\pm\!\sigma$)
& $N$ \\
\midrule
$W_q$
& $0.333\pm0.080$ & $0.224\pm0.070$ & $0.224\pm0.069$
& $0.450\pm0.100$ & $0.316\pm0.089$ & $0.317\pm0.090$
& $0.511\pm0.110$ & $0.368\pm0.100$ & $0.368\pm0.100$ & 28 \\
$W_k$
& $0.369\pm0.060$ & $0.216\pm0.048$ & $0.216\pm0.048$
& $0.554\pm0.079$ & $0.338\pm0.055$ & $0.338\pm0.055$
& $0.637\pm0.078$ & $0.399\pm0.058$ & $0.399\pm0.058$ & 28 \\
$W_v$
& $0.680\pm0.094$ & $0.476\pm0.120$ & $0.475\pm0.119$
& $0.829\pm0.085$ & $0.607\pm0.135$ & $0.606\pm0.135$
& $0.882\pm0.065$ & $0.681\pm0.111$ & $0.677\pm0.114$ & 28 \\
$W_o$
& $0.476\pm0.097$ & $0.378\pm0.081$ & $0.380\pm0.081$
& $0.623\pm0.122$ & $0.515\pm0.107$ & $0.515\pm0.107$
& $0.690\pm0.133$ & $0.580\pm0.120$ & $0.579\pm0.120$ & 28 \\
$W_{up}$
& $0.638\pm0.095$ & $0.503\pm0.119$ & $0.503\pm0.119$
& $0.697\pm0.092$ & $0.561\pm0.124$ & $0.560\pm0.124$
& $0.762\pm0.089$ & $0.623\pm0.129$ & $0.621\pm0.130$ & 28 \\
$W_{down}$
& $0.464\pm0.190$ & $0.293\pm0.190$ & $0.308\pm0.195$
& $0.500\pm0.202$ & $0.321\pm0.207$ & $0.336\pm0.212$
& $0.541\pm0.214$ & $0.355\pm0.226$ & $0.369\pm0.231$ & 28 \\
$W_{gate}$
& $0.341\pm0.134$ & $0.283\pm0.113$ & $0.283\pm0.116$
& $0.381\pm0.145$ & $0.316\pm0.126$ & $0.315\pm0.129$
& $0.427\pm0.157$ & $0.351\pm0.140$ & $0.350\pm0.143$ & 28 \\
\bottomrule
\end{tabular}%
}
\end{table}

\begin{table}[H]
\caption{Per-operator activation-conditioned output error for Qwen2.5~7B under
TT core4 tensorization with $\alpha=1$. Entries report mean $\mu$ and
standard deviation $\sigma$ over per-block sweep values. Lower is better.}
\label{tab:app_qwen_ttcore4_output}
\centering
\footnotesize
\setlength{\tabcolsep}{3.2pt}
\renewcommand{\arraystretch}{1.12}
\resizebox{\textwidth}{!}{%
\begin{tabular}{lccc ccc ccc c}
\toprule
& \multicolumn{3}{c}{$2\times$ compression}
& \multicolumn{3}{c}{$4\times$ compression}
& \multicolumn{3}{c}{$6\times$ compression}
& \\
\cmidrule(lr){2-4}\cmidrule(lr){5-7}\cmidrule(lr){8-10}
Matrix
& Vanilla ($\mu\!\pm\!\sigma$) & Std ($\mu\!\pm\!\sigma$) & MeanAbs ($\mu\!\pm\!\sigma$)
& Vanilla ($\mu\!\pm\!\sigma$) & Std ($\mu\!\pm\!\sigma$) & MeanAbs ($\mu\!\pm\!\sigma$)
& Vanilla ($\mu\!\pm\!\sigma$) & Std ($\mu\!\pm\!\sigma$) & MeanAbs ($\mu\!\pm\!\sigma$)
& $N$ \\
\midrule
$W_q$
& $0.487\pm0.066$ & $0.372\pm0.070$ & $0.373\pm0.072$
& $0.668\pm0.067$ & $0.535\pm0.088$ & $0.533\pm0.090$
& $0.745\pm0.067$ & $0.613\pm0.094$ & $0.608\pm0.095$ & 28 \\
$W_k$
& $0.458\pm0.034$ & $0.364\pm0.030$ & $0.362\pm0.030$
& $0.668\pm0.037$ & $0.557\pm0.040$ & $0.553\pm0.039$
& $0.760\pm0.032$ & $0.651\pm0.042$ & $0.646\pm0.039$ & 28 \\
$W_v$
& $0.607\pm0.040$ & $0.501\pm0.048$ & $0.501\pm0.050$
& $0.794\pm0.031$ & $0.696\pm0.058$ & $0.695\pm0.060$
& $0.862\pm0.025$ & $0.775\pm0.063$ & $0.774\pm0.064$ & 28 \\
$W_o$
& $0.529\pm0.090$ & $0.438\pm0.092$ & $0.443\pm0.096$
& $0.690\pm0.110$ & $0.600\pm0.119$ & $0.598\pm0.124$
& $0.756\pm0.118$ & $0.672\pm0.131$ & $0.666\pm0.135$ & 28 \\
$W_{up}$
& $0.661\pm0.095$ & $0.498\pm0.079$ & $0.503\pm0.081$
& $0.831\pm0.072$ & $0.668\pm0.090$ & $0.670\pm0.095$
& $0.887\pm0.055$ & $0.734\pm0.092$ & $0.734\pm0.098$ & 28 \\
$W_{down}$
& $0.429\pm0.119$ & $0.320\pm0.177$ & $0.321\pm0.178$
& $0.608\pm0.135$ & $0.444\pm0.230$ & $0.443\pm0.232$
& $0.685\pm0.135$ & $0.495\pm0.249$ & $0.493\pm0.249$ & 28 \\
$W_{gate}$
& $0.375\pm0.118$ & $0.289\pm0.115$ & $0.292\pm0.117$
& $0.499\pm0.157$ & $0.391\pm0.151$ & $0.393\pm0.152$
& $0.546\pm0.172$ & $0.433\pm0.165$ & $0.435\pm0.165$ & 28 \\
\bottomrule
\end{tabular}%
}
\end{table}

\FloatBarrier


\FloatBarrier
\clearpage

\end{document}